\documentclass[]{article}

\usepackage{arxiv}

\usepackage{microtype}
\usepackage{graphicx}
\usepackage{subfigure}
\usepackage{booktabs} 
\usepackage{bm}
\usepackage{amsmath}
\usepackage{amsfonts}
\usepackage{amssymb}
\usepackage{color}
\usepackage{pdfpages}
\usepackage{float}
\usepackage{natbib}
\usepackage{algorithm}
\usepackage{algorithmic}
\usepackage{framed}
\usepackage{enumitem}
\usepackage{cancel}
\usepackage{changes}
\usepackage{url}
\usepackage[export]{adjustbox}
\setcitestyle{authoryear,open={(},close={)}}

\def\A{{\mathbf A}}
\def\B{{\mathbf B}}

\def\K{{\mathbf K}}

\def\S{{\mathbf S}}

\def\L{{\mathbf L}}

\def\x{{\mathbf x}}

\def\z{{\mathbf z}}

\def\u{{\mathbf u}}

\def\f{{\mathbf f}}

\def\0{{\mathbf 0}}

\newcommand{\qnew}{{q_{\text{new}}}}

\newcommand{\psiold}{{\bm{\psi}_{\text{old}}}}
\newcommand{\psinew}{{\bm{\psi}_{\text{new}}}}
\newcommand{\uold}{{\u_{\text{old}}}}
\newcommand{\unew}{{\u_{\text{new}}}}

\newcommand{\fnew}{{\f_{\text{new}}}}

\newcommand{\Rbb}{\mathbb{R}}

\newcommand{\Zcal}{\mathcal{Z}}
\newcommand{\Fcal}{\mathcal{F}}
\newcommand{\Ucal}{\mathcal{U}}

\newcommand{\Ocal}{\mathcal{O}}
\newcommand{\Ncal}{\mathcal{N}}
\newcommand{\Dcal}{\mathcal{D}}
\newcommand{\Xcal}{\mathcal{X}}
\newcommand{\Ycal}{\mathcal{Y}}
\newcommand{\Lcal}{\mathcal{L}}

\newcommand{\xc}{\bm{x}}
\newcommand{\yc}{\bm{y}}
\newcommand{\zc}{\bm{z}}

\newcommand{\xcnew}{{\xc_{\text{new}}}}
\newcommand{\ycnew}{{\yc_{\text{new}}}}

\newcommand{\xcold}{{\xc_{\text{old}}}}
\newcommand{\ycold}{{\yc_{\text{old}}}}

\newcommand{\phiold}{{\bm{\phi}_{\text{old}}}}
\newcommand{\phinew}{{\bm{\phi}_{\text{new}}}}

\title{Continual Multi-task Gaussian Processes}
   
\author{%
  	\hspace*{0.35cm}  Pablo Moreno-Mu\~noz$^{1}$ \hspace*{1.0cm} Antonio Art\'es-Rodr\'iguez$^{1}$ \hspace*{1.0cm} Mauricio A. \'Alvarez$^2$ \\
  	$^1$Dept. of Signal Theory and Communications, Universidad Carlos III de Madrid, Spain\\
  	$^2$Dept. of Computer Science, University of Sheffield, UK\\
  	\texttt{\{pmoreno,antonio\}@tsc.uc3m.es},~~\texttt{mauricio.alvarez@sheffield.ac.uk}\\
}

\begin{document}


\maketitle

\begin{abstract}
We address the problem of continual learning in multi-task Gaussian process (GP) models for handling sequential input-output observations. Our approach extends the existing prior-posterior recursion of online Bayesian inference, i.e.\ past posterior discoveries become future prior beliefs, to the infinite functional space setting of GP. For a reason of scalability, we introduce variational inference together with an sparse approximation based on inducing inputs. 
As a consequence, we obtain tractable continual lower-bounds where two novel Kullback-Leibler (KL) divergences intervene in a natural way. 
The key technical property of our method is the recursive reconstruction of conditional GP priors conditioned on the variational parameters learned so far. To achieve this goal, we introduce a novel factorization of past variational distributions, where the predictive GP equation propagates the posterior uncertainty forward.
We then demonstrate that it is possible to derive GP models over many types of sequential observations, either discrete or continuous and amenable to stochastic optimization.
The continual inference approach is also applicable to scenarios where potential multi-channel or heterogeneous observations might appear. Extensive experiments demonstrate that the method is fully scalable, shows a reliable performance and is robust to uncertainty error propagation over a plenty of synthetic and real-world datasets.
\end{abstract}

\section{Introduction}
\label{introduction}

A remarkable evidence of how necessary real-time adaptation is for machine learning can be deduced from multiple medical applications, i.e.\ intensive care unit (ICU) patients or electronic health records (EHR), among others. In such cases, inference methods for probabilistic models typically focus on two principal paradigms: i) discovering the latent structure that underlies a sequence of observations and ii) adapting them to new incoming data.
Out of the medical framework, we often encounter situations where we want to solve multiple tasks that evolve over time, potential examples are signal processing, control, econometrics or even spatio-temporal demographics. 

The resurgence of  interest on probabilistic adaptative methods shows us that, the better the model is adapted to such time evolving behavior, the easier its applicability on real-world problems is. Among the adaptive approaches that we may consider, in this paper we focus on continual ones. Particularly, continual learning, also known as life-long learning, is a very general family of \textit{online} learning methods whose principal properties are the adaptation to non i.i.d.\ data, characterization of tasks that evolve over time and capture of new emergent tasks previously unseen by the model itself.

Gaussian process (GP) models \citep{rasmussen2006gaussian} are not excluded from this necessity of real-time adaptation. Despite their extended use in temporal applications, recursively updating the parameters without revisiting training samples is not trivial. Particularly in such models, the difficulty is double. First, the estimation of non-linear latent functions is constrained by the same principles of \textit{online} Bayesian learning, that is, how to re-introduce  former posterior discoveries as new prior beliefs. Secondly, due to GP priors are based on the construction of covariance matrices via kernel functions, incrementally adapting such matrices to new incoming samples requires expensive ways of matrix completion or even unfeasible inversions when large-scale data is observed.

However, there has been a noticeable effort on adapting GP models for sequential input-output observations over the past decades. As standard Gaussian regression scenarios are usually accompanied by tractable solutions, preliminary works focused exclusively on the iterative counterpart. In particular, this paradigm attracted significant attention since seminal works by \citet{csato2002sparse} and \citet{girard2003gaussian} presented the two preliminar alternatives to perform online predictions using GPs. The first one proposed an online regression model where variational inference is used within moment matching to fit sequential posterior distributions from one single recent sample. In the second case, motivated by one-step ahead predictions, they incorporate an additive input in an equivalent state-space model, which consists of a mapping over the last few observed outputs, $L$ steps back.

Besides initial approaches to \emph{online} GPs, other recent works have also addressed the continual learning problem. For example, sequential rank-one updates of a locally trained GP were proposed in \citet{nguyen2008local} or even label ranking of data points for an inclusion-deletion strategy in an active training set. The GP is learned by Expectation-Propagation (EP) as in \citet{henao2010pass}. Also for the single-output GP case, but closer to the scalable framework presented in this paper, 
we find that the stochastic gradient descent method in \citet{hensman2013gaussian} for Gaussian regression and \citet{hensman2015scalable} for classification, is applicable to online settings but considering ever-increasing datasets, which \emph{a priori} may be problematic. Another recent example is the semi-described (missing inputs) and semi-supervised (missing outputs) GP learning model in \citet{damianou2015semi}, where a forecasting regression problem is seen as a semi-described model where predictions are obtained iteratively in an auto-regressive manner.

In terms of scalability for single-output GP models, both \citet{cheng2016incremental} and \citet{bui2017streaming} extended online learning methods and uncertainty propagation to the popular variational inference setup of sparse GP approximations. They used a novel Kullback-Leibler (KL) divergence  that constrains the new fitted distribution w.r.t.\ the one in the previous instant. While the first work is only related to univariate Gaussian regression problems, the last reference has the additional advantage of accepting limited non-Gaussian likelihoods as well as it is able to include $\alpha$-divergences for more general inference, whose theoretical bounds are analysed in \citet{nguyen2017online}. 

An exception to the previous works is \citet{solin2018infinite}, which instead of employing sparse methods, they use the approximate Markovian structure of Gaussian processes to reformulate the problem as a state-space model. Within this framework, the complexity is reduced from cubic to linear cost in the number of observations, but still stays unfeasible w.r.t.\ the number of states. Introducing a fast EP inference scheme helps to overcome this issue and additionally, the model is able to perform online learning of kernel hyperparameters as well as dealing with non-Gaussian likelihoods. 

Moreover, if we pay attention to the treatment of non-stationary properties, we see that most approaches assume a perpetual latent function behavior which we aim to discover adaptively. In contrast to this assumption, \citet{zhang2019sequential} recently introduced mixtures of GP experts \citep{rasmussen2002infinite} within sequential Monte Carlo (SMC) inference that addresses the variability of such latent functions along time. It is worthy to mention that \citet{solin2018infinite} is also a potential solution for non-stationary structure of models, but using a different approach.

In our paper, we are focused in the general problem of streaming data modelling where samples can be observed as an irregular sequence of batches, one-sample steps or even the case where the complete set of input-output observations is available. Sequential data is not restricted to be i.i.d.\ conditioned to the given model. Additionally, we assume that our dataset might be also high-dimensional and its adaption to non-Gaussian likelihoods is a strict requirement. Similarly to \citet{bui2017streaming}, our model is fitted to the aforementioned constraints, where scalability is addressed through sparse approximations and we use variational inference \citep{titsias2009variational}, which is the standard practice in modern GPs.

Regarding multi-output Gaussian process (MOGP) models, we see that there have been few attempts to extend them to the continual learning scenario. For instance, \citet{cheng2017sparse} contributes to real-time monitoring of patients via structured kernels inside a MOGP model. However, they update the hyperparameters in real-time using momentum methods with a sliding window, rather than discovering the posterior distribution over the latent functions in an online manner. One exception is \citet{yang2018online}, since they derive a variational lower bound for multiple online regression. It is worthy to mention that this is the most closely related work to our multi-output extension, with the important difference that non-Gaussian likelihoods are not considered and neither a variational update of hyperparameters. In contrast to our approach, they use particle filtering given that the model is constrained by a fixed number of inducing-points in the sparse approximation.

Our main contribution in this paper is to provide a novel approach that extends the existing posterior-prior recursion of online Bayesian inference, to the infinite functional space setting of GP models. The key principle in our model is the use of the conditional GP predictive distribution to build a novel implicit prior expression where past posterior discoveries are propagated forward. In addition, we introduce this solution with variational inference for sparse approximations, which avoids any form of data revisiting. The entire model is amenable to stochastic optimization, letting us consider any irregular form in the sequential observation process. Another detail is that the continual learning method is fully applicable to the multi-channel framework, that is, to multi-output Gaussian process models. 

Importantly, the ability of readapting conditional GP priors w.r.t.\ the previous inferred variational distribution is feasible under non-Gaussian likelihoods in the output observations. As non-Gaussian likelihoods are also permitted in the multi-task setup, the continual GP model is useful for heterogeneous problems \citep{morenomunoz2018}. This is the case of several channels for which the outputs are a mix of  continuous, categorical, binary or discrete variables. We also consider asymmetric cases where the observation process of data is not synchronous between channels. Finally, the Python implementation is publicly available with the especial advantage of being easily adapted to multi-task and heterogeneous likelihood problems.

This paper is divided in two main sections that are organized as follows. In Section \ref{sec:continual_gp}, we introduce the sequential data formulation for single-output GPs, that is valid either for univariate regression and classification problems. We then review the deployment of continual variational inference over the sparse GP approximation, where the past data revisiting issue is noticeable. Moreover, we present the recurrent conditional prior reconstruction based on online Bayesian learning that is later used in the definition of our continual lower-bounds. In Section 3, we extend the sequential model for accepting multiple output settings. Particularly, we derive stochastic variational inference for sparse multi-output GPs that follows the same continual learning mechanism but amenable for heterogeneous likelihood models and asymmetric channels setups. Finally, in Section 4, we study the performance of our scalable method on several experiments with synthetic and real-world datasets for both regression and classification tasks.

\section{Continual Gaussian Processes}

\label{sec:continual_gp}

Consider supervised learning scenarios where pairs of input-output data $\Dcal = \{\xc_n, y_n\}^N_{n=1}$ are observed in a sequential manner, with $\xc_n \in \mathbb{R}^p$ and outputs $y_n$ being either continuous or discrete. We assume the sequential observation process to be a finite stream of smaller subsets or batches, such that $\Dcal = \{\Dcal_1, \Dcal_2, \dots, \Dcal_{T}\}$. Additionally, each $t$-th batch, $\Dcal_t = \{\xc_n, y_n\}^{N_t}_{n=1}$, may have an irregular size, that is, different length per batch of data and $N_t < N$ in all cases. From the GP perspective, we consider that every output sample is generated as $y_n \sim p(y_n|f_n)$, where $f_n$ is a non-linear function evaluation $f(\xc_n)$. Here, the latent function $f$ that parameterizes the likelihood model is drawn from a prior $f\sim \mathcal{GP}(0, k(\cdot, \cdot))$, where $k(\cdot, \cdot)$ can be any valid covariance function or kernel, and the zero-mean is assumed for simplicity.

Since we do not know when the next subset $\Dcal_t$ arrives at each time-step, the waiting time and memory allocation resources cannot be estimated \textit{a priori}, mainly due to the size of the batches is being irregular and unknown. Based on \citet{bui2017streaming}, we assume that receiving the entire sequence of data and computing the posterior distribution $p(f|\Dcal)$ is unfeasible and extremely high-time demanding. As alternative, we consider continual learning approaches, which refer to the ability of adapting models in an \textit{online} fashion when data samples are not i.i.d. and updating their parameters without re-observing the entire data sequence. 

In what follows, we will use the notation $\Dcal = \{\Dcal_{\text{old}}, \Dcal_{\text{new}}\}$, where $\mathcal{D}_{\text{old}} = \{\xcold, \ycold\}$ refers to all observations seen so far and the partition $\Dcal_{\text{new}}= \{\xcnew, \ycnew\}$ represents the smaller subset of \textit{new} incoming samples. For this construction, note that if $\Dcal_t$ arrives at a given time, the old data correspond to $\Dcal_{\text{old}}=\{\Dcal_1,\cdots, \Dcal_{t-1}\}$ while $\Dcal_{\text{new}}=\Dcal_t$. This results in an ever-increasing dataset $\Dcal_{\text{old}}$ that is recursively evaluated.

\subsection{Sparse approximations for sequential data}

Exact inference in GP models is widely known for its $\mathcal{O}(N^3)$ complexity for training and $\mathcal{O}(N^2)$ per test prediction. Given the previously described model, the computational effort for learning under such sequential observations could be even more intensive, with a recurrent cost $\mathcal{O}(N_1^3), \mathcal{O}((N_1 + N_2)^3) ,\dots, \mathcal{O}(N^3)$.  In order to sidestep that prohibitive complexity, we introduce \textit{auxiliary} variables also known as \textit{inducing inputs} \citep{snelson2006sparse}. The auxiliary variables serve as an optimal subset of pseudo-observations that summarize the data, reducing the cost of the learning process.

We start by defining the set of inducing inputs $\Zcal = \{\zc_m\}^M_{m=1}$, where $\zc_m \in \mathbb{R}^p$ take values in the same space as $\xc_n$. Moreover, we denote the inducing variables $\u = [u_1, \dots, u_M]^\top$ as the vector of output function evaluations, where $u_m = f(\zc_m)$. Under a construction of this form, the joint distribution $p(y_n, \f_n, \u)$, simplified for a single output sample $y_n$, factorises as

\begin{equation}
	p(y_n, f_n, \u) = p(y_n| f_n)p(f_n, \u) = p(y_n| f_n)p(f_n| \u)p(\u) ,
\end{equation}

where $p(y_n| f_n)$ can be any valid likelihood model and $p(f_n| \u)$, $p(\u)$ are conditional and marginal GP priors respectively. Similarly to the formulation of vectors $\u$, we consider $\f = [f_1, \dots, f_N]^\top$ to be the vector of output function evaluations. 

In practice, obtaining closed-form posterior distributions over both $\f$ and $\u$ is difficult and in many cases, impossible. The problem is generally solved via variational methods, formally denoted with approximations of the form $q(\f, \u) \approx p(\f,\u | \Dcal)$. Following the same derivation of \citet{titsias2009variational}, we assume that the auxiliary distribution $q$ factorises as $q(\f,\u) = p(\f|\u)q(\u)$, reducing the problem to learn a single distribution $q(\u)$ that we assume to be Gaussian.

Importantly, we condition every observed output $y_n$ to the infinite-dimensional function space $f$ similarly to \citet{bui2017unifying}, having $p(y_n| f)$ instead. As a consequence, every variable $f$ will correspond to an infinitely large number of function evaluations, i.e.\ the entire domain $\mathbb{R}^p$, including the input values in $\Zcal$. It will play a key role in the development of the continual inference mechanism later on these lines.\footnote{Infinite dimensional integrals related to $f$ get reduced via properties of Gaussian marginals. The lower bound equation is still tractable. The complete details are included in the Appendix.}

When using variational inference (VI) methods for \textit{sparse} GP models, the common approach is to fit some parameters $\bm{\phi}$ of the auxiliary distribution $q(\u|\bm{\phi})$ by maximizing a lower bound $\Lcal$ on the log-marginal likelihood of the dataset $\log p(\Dcal)$. In the GP literature, this marginal distribution is often rewritten as $\log p(\yc)$ and in our case, we may express it also as $\log p(\ycold, \ycnew)$. From a VI perspective, the log-marginal distribution of the sequential dataset can be decomposed as 
\begin{equation}
	\label{eq:log_marginal}
	\log p(\ycold, \ycnew) = \log \int p(\ycold, \ycnew|f)p(f)df.
\end{equation}

Suppose now that both $\ycold$ and $\ycnew$ are non i.i.d.\ but conditioned to the whole function space $f$, allowing us to apply \textit{conditional independence} (CI). That is, it leads us to obtain the factorized likelihood $p(\ycold, \ycnew|f) = p(\ycold|f)p(\ycnew|f)$ as in \citet{bui2017streaming}, with two separate terms between the old and new data. Then, any standard lower bound $\Lcal$ that we want to build from Eq.  \eqref{eq:log_marginal} would require to evaluate expectations of the form $\mathbb{E}_{q(f)}[ \log p(\ycold, \ycnew|f)]$, where $q(f) = \int p(f|\u)q(\u|\bm{\phi})d\u$ as in the uncollapsed version of the bound \citep{lazaro2011variational,hensman2012fast}. Notice that the evaluation of the expectations is critical due to the difference of size between $\ycold$ and $\ycnew$ might be huge, i.e. millions of samples vs. hundreds respectively. This fact results in very long time computations for re-training with a few more recent observations included in the model, mainly due to the size of the likelihood term $p(\ycold|f)$.

\subsection{Recurrent prior reconstruction}

A meaningful solution for avoiding the sequential evaluation of ever-increasing datasets is approximating old likelihood terms $p(\ycold|f)$ using the previous inferred (joint) variational distribution $q(f|\phiold)$ at each time-step. This idea was first introduced in \citet{bui2017streaming} by means of the Bayes rule, such that

\begin{equation}
	q(f|\phiold) \approx p(f|\ycold, \xcold) \propto p(f)p(\ycold|f),
\end{equation}

where the equality can be inverted to give a proportional estimate of the form

\begin{equation}
\label{eq:lik_approx}
	p(\ycold|f) \approx \frac{q(f|\phiold)}{p(f)}.
\end{equation}

Having the recursive approximation in Eq. \eqref{eq:lik_approx} for old likelihood terms, we can use it to build lower bounds $\Lcal$ where data re-visiting is avoided. Under this strategy, the variational distribution $q(f|\phiold)$ usually factorises according to $p(f_{\neq \u}| \u, \phiold)q(\u|\phiold)$, where $f = \{f_{\neq \u} \cup \u\}$. The main problem that we encounter here is on re-using distributions $q(\u|\phiold)$ estimated over a fixed number of inducing-points $\Zcal_\text{old}$. If for example, the model requires a different subset of inducing inputs $\Zcal_\text{new}$, the previous posterior distribution could not be introduced directly. This is what we will refer as the \textit{explicit} variational distribution issue. Particularly, when we directly introduce Eq. \eqref{eq:lik_approx} in our target lower bound $\Lcal$, what we are doing is to recurrently introduce a summary of our data, through the inducing-points $\u$ and their parameters $\phiold$. In terms of rigorous continual learning, this is another way of revisiting past observed data and forces the GP model to concatenate old and new subsets $\u$, something that can be undesired for certain tasks, i.e.\ high-dimensional input problems.

\subsubsection*{Continual GP prior}

Inspired on online Bayesian inference methods, where \textit{past} posterior distributions are usually taken as \textit{future} priors, our main goal is to reconstruct the GP prior conditioned on the given parameters $\phiold$. The particular construction is as follows. We take the posterior predictive distribution from GP models. It usually is obtained by marginalising the posterior probabilities $p(f|\Dcal)$ given the conditional distribution at test inputs $p(f_*|f)$, whose output values $y_{*}$ we aim to predict. 

Typically,  the predictive distribution takes the form $p(f_*|\Dcal) = \int p(f_*|\u)p(\u|\Dcal)d\u$ when it is applied via sparse approximations.  This posterior predictive formulation is the key idea for recurrently building \textit{continual} GP priors, that is, a new \textit{implicit} distribution at each time step, where all the estimated parameters intervene. For its derivation, we take the appendix A.2 of \citet{alvarez2009variational} as our starting point. Thus, we have a conditional prior of the form

\begin{equation}
\label{eq:cond}
	p(u_*|\u) = \mathcal{N}(u_*|k_{*\u}\K^{-1}_{\u\u}\u, k_{**} - k_{*\u}\K^{-1}_{\u\u}k_{*\u}^\top),
\end{equation}

where  $u_*$ refers to function evaluations $f(\cdot)$ on any arbitrary input-vector $\Zcal_{*}$ that we may consider. Here, the covariance matrix corresponds to $\K_{\u\u} \in \mathbb{R}^{M\times M}$, with entries $k(\zc_i, \zc_j)$ as $\zc_i, \zc_j \in \Zcal_{\text{old}}$ and $k_{*\u} = [k(\cdot, \zc_1), \cdots, k(\cdot, \zc_M)]^\top$. In a similar manner, $k_{**} = k(\cdot, \cdot)$ as in the kernel function of any GP prior. Having the conditional distribution in Eq. \eqref{eq:cond}, which combines both explicit and implicit covariance function constructions, we may use the expectations from the variational distribution $q(\u|\phiold)$ to make the conditional GP prior behave as the former posterior indicates. The process results in a novel \textit{continual} distribution, formally denoted $\widetilde{q}(u_*|\phiold)$, that we obtain as

\begin{equation}
\widetilde{q}(u_*|\phiold) \approx \int p(u_*|\u) q(\u|\phiold)d\u.
\end{equation}

Additionally, if we assume that $q(\u|\phiold) = \Ncal(\u|\bm{\mu}_\text{old}, \S_\text{old})$, then our variational parameters becomes $\phiold = \{\bm{\mu}_\text{old}, \S_\text{old}\}$. Then, the previous expression leads us to an updated GP prior. Its form is

\begin{equation}
\label{eq:continual_prior}
	u_* \sim \mathcal{GP}(k_{*\u}\K^{-1}_{\u\u}\bm{\mu}_\text{old}, k_{**} + k_{*\u}\K^{-1}_{\u\u}(\S_\text{old} - \K_{\u\u})\K^{-1}_{\u\u}k^\top_{*\u}).
\end{equation}

A similar expression is derived in \citet{burt2019rates} where theoretical analysis on sparse GP regression is performed out of the continual learning problem. In particular, the conditional GP prior in Eq.  \eqref{eq:continual_prior} coincides with the approximated posterior process that VI on sparse GP models aims to minimize through the KL divergence \citep{matthews2016sparse}. This result is of particular interest to us, since it provides a closed-form way to introduce Bayesian online learning into GP models, allowing us to naturally avoid any data revisiting, only passing past parameters forward and fixing the posterior-prior recursion.

\subsection{Continual lower-bounds}

Exact posterior inference is still intractable using the previous framework and variational methods are required. However, we are now able to sequentially build lower bounds on the log-marginal likelihood in Eq. \eqref{eq:log_marginal} by only updating from a few recent observations $\Dcal_{\text{new}}$. The continual lower-bound $\Lcal_\mathcal{C}$ is obtained as follows

\begin{equation}
	\label{eq:proto_bound}
	\log p(\ycnew, \ycold) \leq \Lcal_\mathcal{C} \approx  \int q(f|\phinew) \log \frac{p(\ycnew|f)q(f|\phiold)p(f|\psinew)}{q(f|\phinew)p(f|\psiold)}df,
\end{equation}

where $q(f|\phinew)$ is the new variational distribution that we want to update, and $\psiold$ and $\psinew$ are the past and current subsets of hyperparameters involved in the GP prior, respectively. We often use $\bm{\psi}$ to refer both  $\psiold$ and $\psinew$ simultaneously, i.e., $\bm{\psi} = \{\psiold, \psinew\}$. Again, to avoid data revisiting, we have substituted the past likelihood term $p(\ycold|f)$ by its unnormalised approximation, taken from the inverted Bayes rule in Eq. \eqref{eq:lik_approx}. A key difference with respect to \citet{bui2017streaming} appears on the factorisation of our past variational distribution $q(f|\phiold)$. Instead of conditioning on a fixed number of inducing-points $\uold$, we now make use of the continual GP prior in Eq. \eqref{eq:continual_prior}, leading to

\begin{equation}
	q(f|\phiold) = p(f_{\neq u_*}|u_*,\psiold)\widetilde{q}(u_*|\phiold),
\end{equation}

where we extended the factorisation of \citet{titsias2009variational} to accept the entire function space $f$. Moreover, it makes sense to reduce the lower-bound in Eq. \eqref{eq:proto_bound} by critically canceling all conditionals of the form $p(f_{\neq u_*}|u_*)$. Notice that we use $f = \{f_{\neq u_*} \cup u_*\}$ to apply CI. The complete details of this derivation are provided in the Appendix. Then, we obtain the triple-termed bound 

\begin{equation}
\label{eq:int_bound}
\Lcal_\mathcal{C} = \int q(f|\phinew)\log p(\ycnew|f)df - \int q(f|\phinew)\log \frac{q(u_*|\phinew)}{p(u_*|\psinew)}df \nonumber + \int q(f|\phinew)\log \frac{\widetilde{q}(u_*|\phiold)}{p(u_*|\psiold)}df.
\end{equation}

We are now interested in the derivation of a closed-form version of $\Lcal_\mathcal{C}$ that can be evaluated on a specific number of inducing inputs $\Zcal$ rather than on the infinite-dimensional integrals $f$. For that purpose, suppose that our new incoming samples $\Dcal_{\text{new}}$ contain a subset of input values $\xcnew$ whose distance from all the previous ones $\xcold$ is significant. It makes sense to increase the capacity of $\Zcal$ in order to refine the approximated posterior \citep{burt2019rates}. As a consequence, we introduce a new set of inducing variables $\Zcal_\text{new} = \{\zc_m\}^{M_\text{new}}_{m=1}$, where the vector $\unew$ of function evaluations corresponds to $\unew = [u(\zc_1), \cdots, u(\zc_{M_\text{new}}) ]^\top$. Notice that we aim to update the distribution $q(\unew|\phinew) = \mathcal{N}(\unew|\bm{\mu}_\text{new}, \bm{\S}_\text{new})$ where $\phinew = \{\bm{\mu}_\text{new}, \bm{\S}_\text{new}\}$ in this case.

One strategy is that all the distributions that make reference to $u_*$ in $\mathcal{L}_\mathcal{C}$ can be substituted by $\unew$. That is, the former prediction at test-points $\Zcal_*$ are now computed at $\Zcal_\text{new}$. In addition, except for the log-likelihood term in Eq. \eqref{eq:int_bound}, distributions on $f$ may factorise as, for example, $q(f|\phinew) = q(f_{\neq \unew}|\unew, \psinew)p(\unew|\phinew)$, particularly the variational ones. This convenient factorization allows us to use properties of Gaussian marginals, integrating all function values $u_{\neq \unew}$ out of the $\Lcal_{\mathcal{C}}$ bound. Given that, we are able to obtain a closed-form expression of the $\Lcal_{\mathcal{C}}$ bound where three prior and one posterior distributions intervene. Respectively, these terms are: i) the new GP $p(\unew|\psinew)$, ii) the old GP $p(\unew|\psiold)$, iii) the continual GP $\widetilde{q}(\unew|\phiold)$ and iv) the variational posterior $q(\unew|\phinew)$. Then, using the previous expressions we can further simplify $\Lcal_{\mathcal{C}}$ to be

\begin{eqnarray}
\label{eq:continual_bound}
\Lcal_{\mathcal{C}} &=&\mathbb{E}_{q(\fnew)} [\log p(\ycnew|\fnew)] - \text{KL}[q(\unew|\phinew)||p(\unew|\psinew)] \nonumber\\
&+&\text{KL}[q(\unew|\phinew)||p(\unew|\psiold)] - \text{KL}[q(\unew|\phinew)||\widetilde{q}(\unew|\phiold)],
\end{eqnarray}

where $q(\fnew) = \int p(\fnew|\unew)q(\unew|\phinew)d\unew$ as in \citet{saul2016chained}, with $\fnew$ being the vector of output function evaluations $f(\cdot)$ over the inputs $\xcnew$.\footnote{See analytical expression of $q(\fnew)$ in the Appendix.} This functional form of the $\Lcal_{\mathcal{C}}$ bound simplifies the continual learning process to recurrently make the update of parameters
$$\bm{\phi}^{(t+1)}_\text{old}  ~\leftarrow~ \bm{\phi}^{(t)}_\text{new} := \underset{\phinew}{\arg\max}\Big[\Lcal_{\mathcal{C}}\Big(\Dcal^{(t)}_{\text{new}}, \bm{\phi}^{(t)}_\text{old} \Big)\Big].$$

From a practical point of view, when $t=0$ in the expression above, that is, the first time step, we train the model using the bound in \citet{hensman2015scalable} in order to set $\bm{\phi}^{(0)}_\text{new}$. The complete recursive computation of Eq. \eqref{eq:continual_bound} is detailed in Algorithm \ref{alg:single}. Moreover, to learn the variational parameters $\phinew = \{\bm{\mu}_\text{new}, \S_\text{new}\}$, we represent the covariance matrix as $\S_\text{new} = \L_\text{new}\L_\text{new}^\top$. Particularly, we maximise $\Lcal_{\mathcal{C}}$ w.r.t.\ the triangular lower matrix $\L_\text{new}$ to ensure positive definiteness when using unconstrained optimization. In terms of computational effort, the three KL divergence terms in Eq. \eqref{eq:continual_bound} are analytically tractable and of equal dimension (e.g.\ $M_{\text{new}}$). However, depending on the likelihood model considered for $p(\ycnew|\fnew)$, i.e. Gaussian, Bernoulli or Poisson distributed, the expectations could be intractable. For instance, if we observe binary samples $y_n \in [0,1]$, such integrals could be solved via Gaussian-Hermite quadratures, similarly to \citet{hensman2015scalable,saul2016chained}.

The selection of $\Zcal_\text{new}$ is of particular importance for the consistency of the continual learning recursion. Its size, $M_\text{new}$, may vary from the number $M_{\text{old}}$ of previous inducing-points $\Zcal_\text{old}$ without constraints. Notice that, if the incoming batch of samples $\Dcal_{t}$ is determined by some inputs $\xc_{\text{new}}$ that explore unseen regions of $\mathbb{R}^p$, then $\Zcal_\text{new}$ should capture this new corresponding area. However, due to we marginalise former pseudo-observations $\uold$ in Eq. \eqref{eq:continual_prior} for our continual prior construction, either $\Zcal_\text{old}$ and $\Zcal_\text{new}$ are no longer permitted to coincide in any value. If so, the continual bound might not hold, due to a wrong conditioning between variables. However, as we always assume that pseudo inputs $\zc_m$ belong to the real-valued space $\mathbb{R}^{p}$, the problem is generally solved by choosing robust initializations for $\Zcal_\text{new}$. Additional constraints are not needed. 

\begin{algorithm}[]
	\caption{--- \textsc{Continual Gaussian process learning}}
	\label{alg:single}
	\begin{algorithmic}[1]
		\STATE Initialize $\bm{\phi}_\text{new}^{(0)}$ and $\bm{\psi}_\text{new}^{(0)}$ randomly.\\
		\STATE {\bfseries input:} Observe $\Dcal^{(0)}_\text{new}$ 
		\STATE	Maximise $\Lcal\leq\log p(\Dcal^{(0)}_\text{new})$ w.r.t.\ $\{\bm{\phi}_\text{new}^{(0)}, \bm{\psi}_\text{new}^{(0)}\}$. \hfill \textcolor{gray}{$//$ standard variational inference}
		\FOR{$t \in 1,\dots, T$}
		\STATE	Update $\{\bm{\phi}_\text{old}^{(t)}, \bm{\psi}_\text{old}^{(t)}\} \leftarrow \{\bm{\phi}_\text{new}^{(t-1)}, \bm{\psi}_\text{new}^{(t-1)}\}$ \hfill \textcolor{gray}{$//$ past learned parameters become the old ones}
		\STATE Choose initial $\Zcal_\text{new}$ \hfill \textcolor{gray}{$//$ initialization of inducing points}
		\STATE Compute continual GP prior ~$\widetilde{q}(\cdot|\bm{\phi}_\text{old}^{(t)})$ \hfill \textcolor{gray}{$//$ conditional prior reconstruction}
		\STATE {\bfseries input:} Observe $\Dcal^{(t)}_\text{new}$ 
		\STATE	Maximise $\Lcal_\mathcal{C}$ w.r.t.\ $\{\bm{\phi}_\text{new}^{(t)}, \bm{\psi}_\text{new}^{(t)}\}$. \hfill  \textcolor{gray}{$//$ continual variational inference}
		\ENDFOR
	\end{algorithmic}
\end{algorithm}

\subsection{Stochastic continual learning}

Based on \citet{hensman2013gaussian}, we assume that the likelihood model is conditionally independent and fully factorisable across samples,  it holds $p(\yc|\f) = \prod_{n=1}^{N}p(y_n|f_n)$. The likelihood factorisation leads to conditional expectation terms in Eq. \eqref{eq:continual_bound} that are also valid across data observations, allowing us to introduce stochastic variational inference (SVI) methods \citep{hoffman2013stochastic}. In our case, the particular form of the bound $\Lcal_{\mathcal{C}}$ is expressed as
\begin{eqnarray}
\label{eq:stochastic_bound}
\sum_{n=1}^{N_\text{new}}\mathbb{E}_{q(\f_n)} [\log p(y_n|\f_n)] &-& \text{KL}[q(\unew|\phinew)||p(\unew|\psinew)] \nonumber\\&+& 
 \text{KL}[q(\unew|\phinew)||p(\unew|\psiold)] - \text{KL}[q(\unew|\phinew)||\widetilde{q}(\unew|\phiold)].
\end{eqnarray}

So far, under a factorized bound of this form, we are able to combine both continual learning with stochastic optimization, splitting our new incoming subset of data $\Dcal_{\text{new}}$ in smaller \textit{mini-batches} for faster training. Intuitively, it makes the $\Lcal_{\mathcal{C}}$ bound applicable to a wide number of problems, particularly those ones with an extremely asymmetric sequence of observations. That is, if the size of streaming batches is still large for training, we can apply SVI until the next incoming batch will be observed. The combination of SVI with continual learning leads to a \textit{best-of-both-worlds} strategy, since many times stochastic approximations can be also considered for streaming settings \citep{hensman2013gaussian}. In contrast, if the number of new observations goes to the opposite limit,  i.e.\ a reduced number of samples per time-step $t$, then, the stochastic version in Eq. \eqref{eq:stochastic_bound} can be avoided, leading to solutions closer to \citet{solin2018infinite} and Bayesian filtering.

\section{Generalization for Multi-task Models}
\label{sec:continual_mogp}
\begin{figure}[ht]
	\centering
	\includegraphics[width=\textwidth]{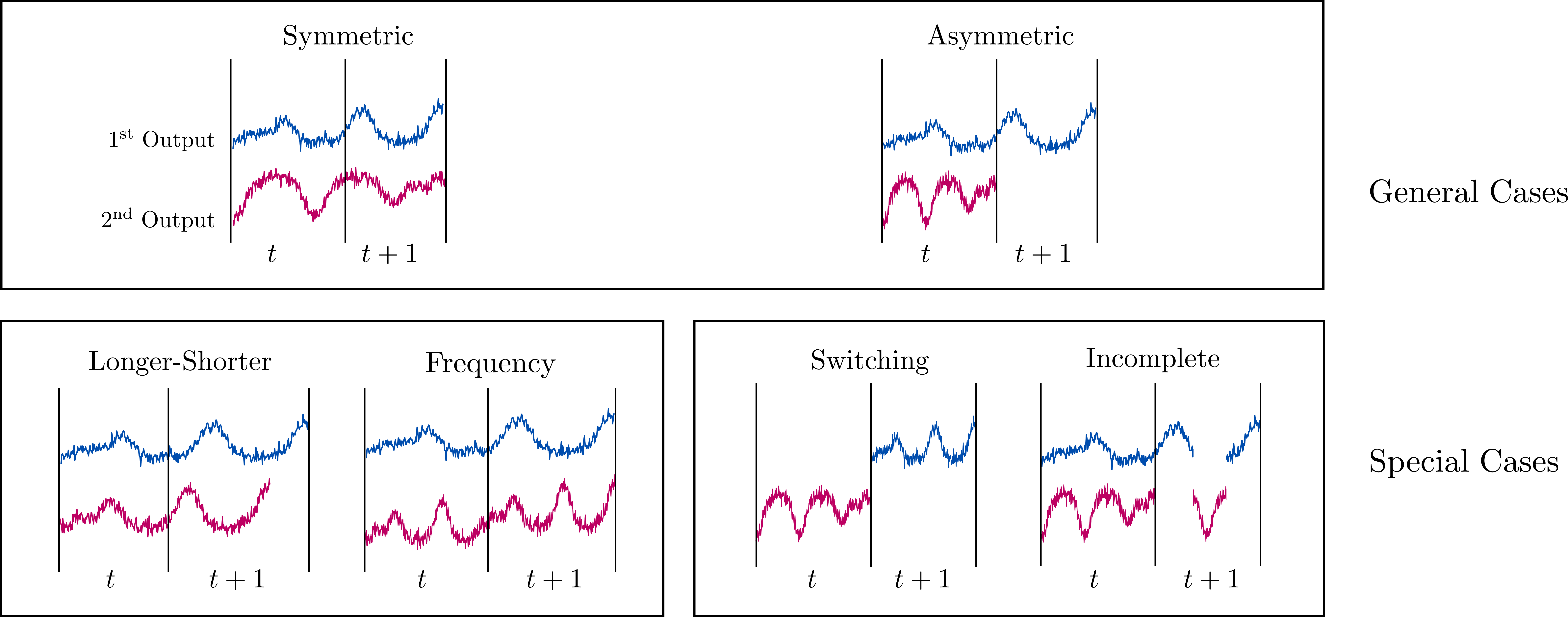}
	\caption{Illustration of the scenarios that two sequences of streaming input-output observations may belong to. \textbf{Upper row.} General cases for the two output channels: symmetric (\textsc{l}) and asymmetric (\textsc{r}) sequential data. \textbf{Lower row.} Special forms of the upper cases: i) one channel is longer at time $t$ (\textsc{l1}), ii) channels have different frequency (\textsc{l2}), iii) switching missing channels (\textsc{r1}) and iv) both outputs sequences are in incomplete (\textsc{r2}). $\textsc{r}=\text{right}$, $\textsc{l}=\text{left}$.}
	\label{fig:mo_cases}
\end{figure}

Regarding the applicability of continual GP priors to high dimensional output settings, we study how to adapt the previous results to sequences of multiple output data. Concretely, we are interested in the generalisation of the continual GP scheme to accept extremely asymmetric cases. For instance, those ones for which, in addition to an unknown stream of  observations, the order of appearance of the multi-output dimensions might be unknown as well. Several cases of both symmetric and asymmetric observation processes are depicted in Figure \ref{fig:mo_cases}.

We begin by considering parallel sequences with different size, formally denoted as \textit{channels}, $\Dcal_d$ with $d \in [1,\dots, D]$. From each $d$-th channel, we sequentially observe batches of input-output data, such that $\Dcal_d = \{\Ycal^{1}_{d},\Ycal^{2}_{d}, \dots, \Ycal^{t}_{d}\}$ where $\Ycal^{t}_{d} = \{y_d(\xc_n)\}^{N^{t}_{d}}_{n=1}$ and $\xc_n \in \mathbb{R}^p$. Notice that here, time steps $t$ are not necessarily aligned across different channels, and its size $N^{t}_{d}$ may also vary. At this point, we initially consider the case for which each $y_d(\xc_n)$ is continuous and Gaussian distributed. The assumption will be relaxed later on this section.

Having a multiple output problem of this type, we want to jointly model it using multi-output Gaussian processes (MOGP). These models generalise the flexible prediction system of GP approaches to the vector-valued random field setup \citep{alvarez2012kernels}. Particularly, it is demonstrated that by exploiting correlations among different streams of outputs, or channels, they are able to improve in the prediction for every $d$-th output. We aim to exploit the idea of correlated outputs in the multi-task sequential framework. However, little work has been done on extending MOGP models to the continual learning scenario. The most closely related works to ours are \citet{cheng2017sparse} and \citet{yang2018online}. Importantly, we are different from \citet{cheng2017sparse} because we allow for continual updates of the MOGP model while they focus on adding structure to the kernel functions. The work by \citet{yang2018online} also derives tractable variational lower bounds based on the sparse approximation, but they do not handle non-Gaussian likelihoods and the learning method uses particle filtering with a fixed number of inducing points. In this section, we present a novel extension to perform continual learning given any MOGP model, independently of the likelihood distributions considered.

\subsection{Multi-parameter GP prior}

The following description of the multi-parameter GP prior is built on the heterogeneous MOGP model \citep{morenomunoz2018}. Based on the single-output model presented above, we begin by defining the set of Gaussian likelihoods for each set of output vector values $\yc_d$ given a channel $\Dcal_d$,  such that $\yc_d = [y_d(\xc_1),y_d(\xc_2), \cdots, y_d(\xc_{N^{t}_d})]^\top.$ We also assume for every batch that its samples are conditionally independent (CI) given the vector of parameter functions $\bm{\theta}_d(\xc)\in \Xcal^{J_d} $, where $\Xcal$ is the specific domain for each parameterisation and $J_d$ is the number of parameters that define the target distribution. In the particular case of standard GP regression,  the set $\bm{\theta}_d(\xc)$ corresponds to the mean parameter $\mu_d(\xc) \in \mathbb{R}$, which is assumed to be a non-linear function $f_d(\xc)$ drawn from a GP prior. This means that we use $J_d = 1$ in this first approach, with $\mu_d(\xc) = f_d(\xc)$ for all outputs. A potential exception would be \textit{linking} several functions together to the same parameter $\theta_{d,j}(\xc)$ as in \citet{saul2016chained}, or \textit{casting} the standard deviation as positive-real valued function $\sigma_d(\xc)$, i.e.\ heteroscedastic GP regression \citep{lazaro2011variational}. Both extensions are applicable to the present approach, but we avoid them for the reason of simplicity in the notation. Our definition for every likelihood distribution of $\yc_d$ is therefore

\begin{equation}
	\label{eq:mo_likelihood}
	p(\yc_d|\bm{\theta}_d(\xc)) = p(\yc_d|\f_d(\xc)) = \Ncal(\yc_d|\f_d(\xc), \sigma^2_d\mathbb{I}),
\end{equation}

where we specify the vector of latent output functions (LOF) as $\f_d(\xc) = [f(\xc_1), f(\xc_2), \cdots, f(\xc_{N_t})]^\top\in \mathbb{R}^{N_t\times 1}$, that here acts as the mean vector function of the aforementioned Gaussian distributions. Importantly, notice that the likelihood noise variances $\sigma_d$ are assumed to be fixed. Hence, if we consider single-output approaches for every channel, we would have $D$ independent priors for each $f_d$ such that $f_d \sim \mathcal{GP}(0,k_d(\cdot,\cdot))$, with $k_d$ being different kernel functions. 

Notice that, since our goal is to build a multi-parameter prior, we correlate all output parameter functions $\Fcal = \{f_d(\xc)\}^D_{d=1}$ together. That is, we jointly model the output channels through the linear model of corregionalisation (LMC) \citep{journel1978mining}. The construction of the multi-output prior is as follows.

Instead of using a single GP prior per $f_d$, we introduce an additional set of independent latent functions (LF) denoted by $\Ucal = \{u_q(\xc)\}^Q_{q=1}$. Moreover, we assume that latent functions $u_q(\xc)$ are linearly combined to produce $D$ LOFs, that is, functions $\Fcal$ that are conditionally independent given $\Ucal$. Then, each latent function is assumed to be drawn from an independent GP prior, such that $u_q(\cdot) \sim \mathcal{GP}(0,k_q(\cdot,\cdot))$, where $k_q$ is any valid covariance function. Under this construction, each function $f_d(\xc)$ is given by
\begin{equation}
	\label{eq:lmc_function}
		f_d(\xc) = \sum_{q=1}^{Q}\sum_{i=1}^{R_q}a^i_{q,d} u^i_q(\xc),
\end{equation}
where coefficients $a^i_{q,d} \in \Rbb$ and all $u^i_q$ are i.i.d.\ realizations from the GP $u_q(\cdot)$. Given $Q$ zero-mean priors, the mean function for $f_d(\xc)$ is set to zero as well. Any cross-covariance matrix between output functions can be built as $k_{f_df_{d'}}(\xc,\xc') = \text{cov}[f_d(\xc), f_d'(\xc')]$, which is equal to $\sum_{q=1}^{Q}b^q_{d,d'}k_q(\xc,\xc')$, where $b^q_{d,d'} = \sum^{R_q}_{i=1}a^i_{q,d} a^i_{q,d'}$. Thus, we obtain the matrix $\B_q\in \mathbb{R}^{D\times D}$, whose entries are $\{b^q_{d,d'}\}^{D,D}_{d=1, d'=1}$. Alternatively, matrices $\B_q$ can be also formulated as $\A_q\A_q^\top$, where $\A_q$ has entries $\{a^i_{q,d}\}^{D,R_q}_{d=1, i=1}$. In this work, we always assume $R_q = 1$, that is, we take a single sample per each independent $q$-th GP prior, reducing coregionalisation matrices to be rank-one. This model is also known in the literature as the \textit{semiparametric latent factor} model \citep{teh2005semiparametric}.  

It is important to remark that besides using LMC as the combination of LFs $u_q(\cdot)$ to get $D$ potential output functions $f_d(\cdot)$, the multi-output model can also accept other valid operators as, for example, \textit{convolutional} processes \citep{alvarez2009sparse} or non-linear combinations with Volterra series \citep{alvarez2019non}. 

\subsection{Sequential multi-output formulation}

Having a multi-parameter GP prior with the aforementioned form, we want to model the sequential observation process properly. Suppose that we expect to observe a high-dimensional dataset $\mathcal{D}=\{\xc_n, \yc_n\}^N_{n=1}$ where we know \textit{a priori} that output vectors $\yc_n \in \Rbb^{D\times 1}$ are composed by $D$ features, such that $\yc_n = [y_1(\xc_n), y_2(\xc_n),\cdots, y_D(\xc_n)]^\top$ with $\xc_n \in \Rbb^p$ as in the single-output scenario. Again, we assume that the data $\mathcal{D}$ will be observed as a flow of smaller batches $\Dcal_1, \Dcal_2, \cdots, \Dcal_t$ with irregular size and unknown arrival time. We also suppose that the pairs of output-input observations are aligned between channels, that is, the streaming setting is equivalent to the single-output case but considering output vectors $\yc_n$ instead of scalars for simplicity in the derivation. Importantly, the multi-output model presented here is also applicable to the case of asymmetric channels (see Figure \ref{fig:mo_cases}), as we will show later on this section.

\textcolor{black}{The generative process of the multi-output samples is as follows. We assume that there exist $Q$ latent functions $\Ucal$ that are linearly combined to produce $D$ latent output functions $\Fcal$ along time, using the LMC formulation. In our MOGP prior, each one of the $\Ucal$ functions is stationary across batches $\Dcal_t$ and their output variables $\yc_n$ follow a probability distribution $p(\yc_n|\f_n) = \prod_{d=1}^{D}p(y_d(\xc_n)|f_d(\xc_n))$. We also define the vector $\f_n = [\f_1^\top, \f_2^\top, \cdots, \f^\top_D]^\top \in \Rbb^{DN_t \times 1}$.} Moreover, we reuse the notation from the single-output case to indicate that our dataset is recursively partitioned, as $\Dcal = \{\Dcal_{\text{old}}, \Dcal_{\text{new}}\}$, where $\Dcal_{\text{new}} = \Dcal_t$ at each time step $t$ and $\Dcal_{\text{old}}$ ever increases.

When training the MOGP model for exact inference, the problem is analogous to the continual GP case. This is, we encounter a recurrent  computational cost that now also includes $D$, the number of outputs, such that $\Ocal(D^3N_1^3), \Ocal(D^3(N_1+N_2)^3), \cdots, \Ocal(D^3N^3)$. Even if we avoid the use of non-Gaussian likelihoods for every output, where exact posterior distributions are intractable, such computational cost is still unfeasible. Therefore, inducing variables are introduced within variational inference for the reason of scalability. Sparse approximation methods have been already used in the context of MOGP \citep{alvarez2009sparse,alvarez2010efficient,morenomunoz2018}. The subtle difference from the single-output case lies on the fact that pseudo-observations are not taken from the output functions $\Fcal$ but from the latent ones $\Ucal$ instead. Consequently, the extra layer that the multi-output GP adds for correlating latent functions, is also used for the sparse approximation, inducing a two-step conditioning on the model. For instance, output functions  values are conditioned to latent functions and latent function vectors are conditioned to the subset of pseudo-observations. \textcolor{black}{Under this setting, we define $Q$ sets of $M_q$ inducing variables, one per function $u_q(\cdot)$, such that $\zc = \{\zc_m\}^{M_q}_{m=1} \in \Rbb^{M_q\times p}$. It is important to mention that these subsets are not restricted to take the same values of $\zc_m$ across dimensions and neither the same size $M_q$.  However, we consider all $M_q$ to be identical and equal to $M$ in this work, for simplicity in the notation. We also denote $\u_q =[u_q(\zc_1), u_q(\zc_2), \cdots, u_q(\zc_{M})]^\top$ as the vector of LF evaluations given the $u_q$ process and  $\u = [\u^\top_1, \u^\top_2, \cdots, \u^\top_Q]^\top \in \Rbb^{QM\times 1}$ for the whole set of functions $\Ucal$. Notice that here, we have the sparse GP notation transformed for the multi-output problem.}

Given $D$ output functions $\Fcal$ and $Q$ latent functions $\Ucal$, we build our joint prior to be $p(\Fcal, \Ucal) = p(\Fcal| \Ucal)p(\Ucal|\bm{\psi})$, where again, we use $\bm{\psi}$ to refer the subset of hyperparameters involved in the MOGP prior. Using the infinite-dimensional approach that we introduced in the single-output case, we can factorize our prior by conditioning on the finite number of inducing points $\u$ as
\begin{equation}
	p(\Ucal|\bm{\psi}) = p(\Ucal_{\neq \u}|\u, \bm{\psi})p(\u|\bm{\psi}),
\end{equation}

where $\Ucal_{\neq\u}$ refers to all latent functions values $\Ucal$ not including $\u$, that is, $\Ucal = \Ucal_{\neq\u} \cup \u$. The prior distribution over $\u$ also factorises across latent functions, as $p(\u|\bm{\psi}) = \prod_{q=1}^{Q}p(\u_q|\bm{\psi})$ with $\u_q \sim \mathcal{N}(\bm{0},\K_q)$ and $\K_q \in \mathbb{R}^{M \times M}$ corresponds to $k_q(\zc_i, \zc_j)$ with entries $\zc_i$, $\zc_j \in \zc$.  The dimension of $\K_q$ always changes within the number of inducing points evaluations, determining the model's maximum complexity. This last detail plays an important role when the input domain is incremental within the appearance of newer observations \citep{burt2019rates}.


Hence, our primary goal is to obtain the posterior distribution $p(\f,\u| \Dcal)$, that we know is analytically intractable under the presence of inducing points and potential non-Gaussian likelihoods. If we consider the variational approach as in \citet{titsias2009variational}, where we can approximate our posterior with an auxiliary Gaussian distribution $q(\cdot, \cdot)$, we may consider the following factorisation as in \citet{alvarez2010efficient}.
\begin{equation*}
p(\f,\u|\Dcal) \approx q(\f,\u) = p(\f|\u)q(\u) 
= \prod_{d=1}^{D}p(\f_{d}|\u)\prod^Q_{q=1}q(\u_q),
\end{equation*}
where we have a product of $Q$ Gaussian distributions, one per latent process, with $q(\mathbf{u}_q) =\mathcal{N}(\u_q|\bm{\mu}_{\u_q}, \S_{\u_q})$ and where the conditional distribution $p(\f_d|\u)$ is given by

\begin{equation*}
p(\f_d|\u)  
= \mathcal{N}\Big(\f_{d}|\K_{\mathbf{f}_{d}\u}\mathbf{K}^{-1}_{\u\u}\u, \mathbf{K}_{\f_{d}\f_{d}}  - \mathbf{K}_{\f_{d}\u}\K^{-1}_{\u\u}\K^{\top}_{\f_{d}\u}  \Big),
\end{equation*}

with $\K_{\f_{d}\u} \in \mathbb{R}^{N\times QM}$ being the cross-covariance matrix obtained by evaluating correlation between $f_d(\x)$ and $u_q(\z)$. We also denote $\K_{\u\u}\in \mathbb{R}^{QM\times QM}$ as the block-diagonal matrix formed by the $\K_q$ matrices.

\subsection{Avoiding revisiting multiple likelihoods}

When using variational inference, we fit the distributions $q(\u_q)$ by maximising a lower bound $\Lcal$ of the log-marginal likelihood $\log p(\Dcal)$. In the MOGP literature, this marginal is also written as $\log p(\yc)$ and in our case, we express it also as $\log p(\ycold, \ycnew)$. Given the previously defined sparse MOGP model, this probability distribution can be decomposed as a double integral
\begin{equation}
\label{eq:marginallik}
\log p(\ycnew, \ycold) = \log \iint p(\ycnew, \ycold|\Fcal)p(\Fcal, \Ucal)d\Fcal d\Ucal,
\end{equation}

where we now consider the finite set of output values $\ycold$ and $\ycnew$ to be conditioned on the set of whole function domains $\Fcal$ as in \citet{bui2017streaming} but for the multiple output case. Due to this assumption, we have a double integration over both $\Fcal$ and $\Ucal$ where we can apply conditional independence in the likelihood term of (\ref{eq:marginallik}). This leads us to obtain $p(\ycnew, \ycold|\Fcal) = p(\ycnew|\Fcal)p(\ycold|\Fcal)$. For simplicity, we will denote both terms as the \textit{new} and \textit{old} likelihoods respectively.

As it was previously mentioned, when dealing with variational inference, any standard lower bound $\Lcal$ over \eqref{eq:marginallik} requires to sequentially evaluate expectations given former log-likelihood terms $\log p(\ycold|\f)$. However, under the assumption of a multi-output GP model, the recurrent evaluation of expectations even worsens. In particular, due to the factorization of LOFs,  it is necessary to compute, at least, $D$ integrals over the dimensions of old data vectors $\ycold$. Notice that each $d$-th  dimension might be characterized by a different likelihood function that we aim to estimate through expected values. Fortunately, the meaningful solution of \citet{bui2017streaming} still yields in our multiple channel setting. We can approximate all probabilities $p(\ycold|\f)$ by means of the Bayes rule. We have that as long as 
\begin{equation}
q(\Fcal, \Ucal) \approx p(\Fcal, \Ucal|\ycold, \xc_{\text{old}}) \propto p(\Fcal, \Ucal)p(\ycold|\Fcal),
\end{equation}

we can invert the Bayes rule equality to obtain an unnormalized estimate of the likelihood term $p(\ycold|\Fcal)$ as
\begin{equation}
\label{eq:likapprox}
p(\ycold|\Fcal) \approx \frac{q(\Fcal, \Ucal)}{p(\Fcal, \Ucal)}.
\end{equation}
Importantly, the two distributions that intervene in the quotient of Eq. $\eqref{eq:likapprox}$ factorize as follows
\begin{align}
&q(\Fcal, \Ucal) = p(\Fcal|\Ucal)p(\Ucal_{\neq \u}|\u, \psiold)\prod_{q=1}^{Q}q(\u_q), \label{eq:var_factor1}\\
&p(\Fcal, \Ucal) = p(\Fcal|\Ucal)p(\Ucal_{\neq \u}|\u, \psiold)\prod_{q=1}^{Q}p(\u_q|\psiold),
\end{align}
where both variational posteriors $q(\cdot)$ and priors $p(\cdot)$ are evaluated over the inducing points given the respective $Q$ latent functions. This fact will make it easier to obtain separated KL divergence terms in the future continual lower bound for multi-task problems. Additionally, if we introduce the aforementioned expression in Eq. $\eqref{eq:likapprox}$ as a sequential estimator of our multiple old likelihood terms given some previous inferred distribution $q(\Fcal, \Ucal|\phiold)$, we can reformulate Eq. \eqref{eq:marginallik} to be

\begin{equation}
\label{eq:approx_marginal}
\log p(\ycnew, \ycold)  \approx \log \iint \frac{p(\ycnew|\Fcal)p(\Fcal, \Ucal)q(\Fcal, \Ucal)}{p(\Fcal, \Ucal)}d\Fcal d\Ucal,
\end{equation}

where both prior distributions $p(\Fcal, \Ucal)$ in the quotient differ given different subsets of hyperparameters, i.e.\ the new $\psinew$ and the former ones $\psiold$. Having an approximated log-marginal distribution of this form, we can build our lower bound $\Lcal \leq \log p(\ycnew, \ycold)$ by means of the Jensen's inequality and without revisiting past samples.

\begin{equation}
\label{eq:approx_marginal}
 \Lcal =  \iint q(\Fcal, \Ucal|\phinew)\log \frac{p(\ycnew|\Fcal)p(\Fcal, \Ucal)q(\Fcal, \Ucal)}{q(\Fcal, \Ucal|\phinew)p(\Fcal, \Ucal)}d\Fcal d\Ucal.
\end{equation}

As previously mentioned on Section \ref{sec:continual_gp}, there is still a problem related to the use of past \emph{explicit} distributions in continual lower bounds $\Lcal$, e.g.\ reusing distributions evaluated over past inducing points might be problematic. This issue remains in the multi-output setup as we have to propagate past inducing points $\uold$ forward, for each latent function, in order to approximate likelihood terms with the expression in Eq. \eqref{eq:var_factor1}. To avoid it, we adapt the continual GP prior idea within the predictive expressions to the multiple output setting.

Consider an arbitrary set of test inducing inputs $\Zcal_*$. Assumming that $p(\u|\Dcal) \approx q(\u)$, the predictive distribution $p(\Ucal_*|\Dcal)$ can be approximated as $\int p(\Ucal_*|\u)q(\u)d\u$, where we used $\Ucal_*$ to denote the LF values taken on $\Zcal_*$. While $q(\u)$ factorises accross the $Q$ latent functions vectors $\u_q$, the conditional multi-output prior $p(\Ucal_*|\u)$ is analogous to the one that we obtained in Eq. \eqref{eq:cond} but having block matrices $\K_q$ instead. This means that we have the same mechanism used to build continual GP priors, that now works similarly but in the latent function layer rather than in the output function one obtained after mixing. As a consequence, for each one of the $q$-th non-linear functions, we will set a continual GP prior of the form $\widetilde{q}(u_*|\phiold) \approx \int p(u_{*}|\u_q) q(\u_q|\phiold)d\u_q$.  Moreover, due to every one of the latent functions has its own independent covariance function, the continual update process is separated as well. In particular, we assume the existence of $Q$ parallel priors of the form
\begin{equation}
\label{eq:mogp_continual_prior}
u_{q,*} \sim \mathcal{GP}(k_{*\u_q}\K^{-1}_{\u_q\u_q}\bm{\mu}_{q,\text{old}}, k_{**} + k_{*\u_q}\K^{-1}_{\u_q\u_q}(\S_{q,\text{old}} - \K_{\u_q\u_q})\K^{-1}_{\u_q\u_q}k^\top_{*\u_q}),
\end{equation}

where $k_{*\u_q} = [k_q(\cdot, \zc_1), \cdots, k_q(\cdot, \zc_{M_q})]^\top$ refers to the values taken on the corresponding kernel constructor. The development of the multi-output version of the continual lower bound is now feasible. First, we use the predictive prior to factorize the expression in Eq. \eqref{eq:var_factor1} as $q(\Fcal, \Ucal) = p(\Fcal|\Ucal)p(\Ucal_{\neq \u}|u_{q,*}, \psiold)\prod_{q=1}^{Q}q(u_{q,*})$, where, for instance, we can set $u_{q,*} = \u_{q,\text{new}}$ to make the prior-posterior recursion available. Hence, we can further simplify $\Lcal$ by means of the continual predictive prior and Gaussian marginals properties to be

\begin{align}
\Lcal &= \iint q(\Fcal, \Ucal|\phinew) \log \frac{p(\ycnew|\Fcal)p(\unew|\psinew)}{q(\unew|\phinew)}d\Fcal d\Ucal  + \iint q(\Fcal, \Ucal) \log \frac{q(\unew|\phiold)}{p(\unew|\psiold)}d\Fcal d\Ucal.
\end{align}

This expression can also be rewritten in a more recognizable way like
\begin{align}
\label{eq:final}
\mathcal{L} =& \sum_{d=1}^{D}\mathbb{E}_{q(\f_{d, \text{new}})}\left[\log p(\yc_{d, \text{new}}|\f_{d, \text{new}})\right] -\sum_{q=1}^{Q}\text{KL}\left[q(\u_{q,\text{new}}|\phinew)||p(\u_{q,\text{new}}|\psinew)\right] \nonumber\\
+& \sum_{q=1}^{Q} \text{KL} \left[\qnew(\u_{q,\text{new}}|\phinew)||p(\u_{q,\text{new}}|\psiold)\right] - \sum_{q=1}^{Q} \text{KL}\left[q(\u_{q,\text{new}}|\phinew)||q(\u_{q,\text{new}}|\phiold)\right],
\end{align}
where $q(\f_{d, \text{new}}) = \mathbb{E}_{q(\unew|\phinew)}[p(\f_{d, \text{new}}|\unew)]$ is the approximate marginal posterior for every $\f_{d, \text{new}} = f_d(\xcnew)$ that can be obtained analytically via
\begin{align*}
q(\f_{d, \text{new}}) =
\mathcal{N}(\f_{d, \text{new}}&|\mathbf{K}_{\f_{d, \text{new}}\unew}\mathbf{K}^{-1}_{\unew\unew}\bm{\mu}_{\unew},\K_{\f_{d, \text{new}}\f_{d, \text{new}}} \\&+ \mathbf{K}_{\f_{d, \text{new}}\unew}\mathbf{K}^{-1}_{\unew\unew}(\S_\unew -\K_{\unew\unew})\mathbf{K}^{-1}_{\unew\unew}\K_{\f_{d, \text{new}}\unew}^{\top}), 
\end{align*} 
where $\bm{\mu}_{\unew} = [\bm{\mu}^\top_{\u_{1,\text{new}}}, \cdots, \bm{\mu}^\top_{\u_{Q,\text{new}}}]$ and $\S_\unew$ is a block matrix whose elements are given by $\K_{\u_{q,\text{new}}}$. The interpretability of the multi-output continual bound in Eq. \eqref{eq:final} is of particular interest in our work. In the single-output case, both expectations and divergence terms refer to the same layer of computation, that is, the one where both observations and output functions $f(\cdot)$ lie and are parameterising the likelihood distribution. However, in the the multi-output setting, the expectation term in Eq.\eqref{eq:final} is focused at the observation counterpart, while the KL regularization terms exclusively affects the layer of the latent functions $\Ucal$. Particularly, the three KL divergences regularise the continual variational inference process that will be updated sequentially if, for instance, the input domain increases along time. In constrast,  we have $D$ expectation terms on a different layer, which are \textit{invisible} to the continual learning mechanism due to they are only evaluated conditioned to the most recently learned parameters. This property makes the method applicable to asymmetric scenarios or where, for instance, one of the channels might be unobserved after some time step.

\subsection{Stochastic updating and heterogeneous likelihoods}

The present approach is also valid when the continual lower bound in Eq. \eqref{eq:final} factorises across data observations. The expectation term $\mathbb{E}_{q(\f_{d, \text{new}})}\left[\log p(\yc_{d, \text{new}}|\f_{d, \text{new}})\right]$ is there expressed as a $N$-dimensional sum, amenable for stochastic variational inference \citep{hoffman2013stochastic,hensman2013gaussian,morenomunoz2018} by using small subsets of training samples. The optimization method uses noisy estimates of the global objective gradient at each time step of the sequential process. Similar stochastic updates have been already used in \citet{hensman2013gaussian,hensman2015scalable,saul2016chained,morenomunoz2018}. The scalable bound makes our continual multi-output model applicable to larger datasets, i.e.\ multi-channel patient monitoring signals or ICU time-series, among others.

An important detail to consider is the hyperparameter learning, that is, the sequential update of variables associated to the covariance functions $\{k_q(\cdot, \cdot)\}^Q_{q=1}$ that  have been previously denoted as $\psiold$ and $\psinew$. Due to abrupt changes in the hyperparameters may affect the learning process of $q(\unew|\psinew)$, which is sensitive to amplitude or smoothness, we use several steps of the variational EM algorithm \citep{beal2003variational} within each sequential update. This makes the optimization process more stable through coordinate ascent. In Algorithm \ref{alg:multi}, we present all necessary computations for continually learning the proposed MOGP model. The key difference between Algorithms \ref{alg:single} and \ref{alg:multi} is that the latter one requires $Q$ iterations over the LFs.

Additionally, having presented the previous continual model for multi-output Gaussian Process regression,  we may effortlessly consider the apparition of other non-Gaussian output variables in the sequential dataset $\Dcal$. Besides popular scalable GP methods for dealing with non-Gaussian likelihoods \citep{Dezfouli2015blackboxGP,hensman2015scalable} in both single and multi-output scenarios, we focus in the open problem of \textit{heterogeneous} likelihood models. In this case, each $d$-th output can be either a continuous, categorical, binary or discrete variable, following a different likelihood model.  For this general situation, we can adapt the current construction of the continual lower-bounds to accept heterogeneous MOGP models \citep{morenomunoz2018}. Particularly, we generalise the probability distributions followed by the outputs values $y_d$ in $\Ycal$ to accept any valid combination of likelihood functions. Notice that in the multi-output GP framework, the continual learning mechanism is placed exclusively on the latent function layer. Hence, the appearance of a new likelihood distribution only affects to the $d$-th expectation term present in the l.h.s.\ of Eq. \eqref{eq:final}.

\begin{algorithm}[]
	\caption{--- \textsc{Multi-channel continual GP learning}}
	\label{alg:multi}
	\begin{algorithmic}[1]
		\STATE Initialize $\bm{\phi}_\text{new}^{(0)}$ and $\bm{\psi}_\text{new}^{(0)}$ randomly.\\
		\STATE {\bfseries input:} Observe $\Dcal^{(0)}_\text{new}$ 
		\STATE	Maximise $\Lcal\leq\log p(\Dcal^{(0)}_\text{new})$ w.r.t.\ $\{\bm{\phi}_\text{new}^{(0)}, \bm{\psi}_\text{new}^{(0)}\}$. \hfill \textcolor{gray}{$//$ standard variational inference}
		\FOR{$t \in 1,\dots, T$}
		\STATE	Update $\{\bm{\phi}_\text{old}^{(t)}, \bm{\psi}_\text{old}^{(t)}\} \leftarrow  \{\bm{\phi}_\text{new}^{(t-1)}, \bm{\psi}_\text{new}^{(t-1)}\}$ \hfill \textcolor{gray}{$//$ past learned parameters become the old ones}
		\FOR{$q \in 1,\dots, Q$}
		\STATE {\bfseries input:} Observe $\Dcal^{(t)}_\text{new}$ 
		\STATE Choose initial $\Zcal_\text{new}$ \hfill \textcolor{gray}{$//$ initialization of inducing points}
		\STATE Compute continual GP priors ~$\widetilde{q}(\cdot|\bm{\phi}_\text{old}^{(t)})$ \hfill \textcolor{gray}{$//$ conditional prior reconstruction}
		\ENDFOR
		\STATE	Maximise $\Lcal_\mathcal{C}$ w.r.t.\ $\{\bm{\phi}_\text{new}^{(t)}, \bm{\psi}_\text{new}^{(t)}\}$. \hfill  \textcolor{gray}{$//$ continual variational inference}
		\ENDFOR
	\end{algorithmic}
\end{algorithm}

\section{Experiments}
\label{sec:experiments}

Our experiments in this paper are focused in three main topics that aim to demonstrate the utility and robustness of the approach over both toy and real-world datasets. The three topics are: i) performance of the continual GP model under single-output streaming observations, ii) resistance to propagation errors when reusing variational approximations, including fitting to the appearance of tasks, non-Gaussian data and heterogeneous multi-output settings, iii) applicability to real world problems with multi-dimensional \textit{online} data, potentially configured as asymmetric channels. A particular detail of the aforementioned experiments is that they are organized into several subsections related to single-output regression, classification, multi-channel settings and last, heterogeneous likelihood models. 

For all experiments, we used a modified version of the \textsc{Python} code released within \citet{morenomunoz2018} that presents similar features of scalability and adaptability to multi-output and non-Gaussian data. For the optimization process w.r.t.\ continual lower bounds $\Lcal_{\mathcal{C}}$, we make use of the LBFGS-B algorithm and when the stochastic counterpart is necessary, we considered ADADELTA instead, which is included in the \textit{climin} library. Further details about the general setting of hyperparameters are included in the Appendix. Moreover, our code is publicly available in the repository  \url{github.com/pmorenoz/ContinualGP/} where all the experiments included in this section can be fully reproduced.

\subsection{Continual GP regression} 

In our first subset of experiments, we evaluate the performance of the continual GP approach for the case of single-output scenarios where streaming data is real-valued, assumed Gaussian distributed and we aim to perform sequential non-linear regression. We first setup a toy problem with three different versions in the way of appearance of the incoming samples. We denote them as  i) streaming, ii) overlapping and iii) incremental data. In the first case, we have a sequence of $t=10$ non-overlapping partitions that are recursively delivered to the learning system. Each partition avoids revisiting the previously explored input domain. Secondly, we relax the assumption of non-overlapping partitions of data to consider partially overlapping tasks where parts of the input domain may also be re-visited (not the observations).  The last version of the experiment refers to the same dataset that now is progressively completed within the emergence of new batches. Importantly, we always use a single-output latent function for modeling likelihood parameters $\bm{\theta}$, that is, we avoid solutions similar to the chained GP \citep{saul2016chained}, which could be also applied to the current experiment with continual GPs.

\begin{figure}[] 
	\centering
		\includegraphics[width=0.75\textwidth]{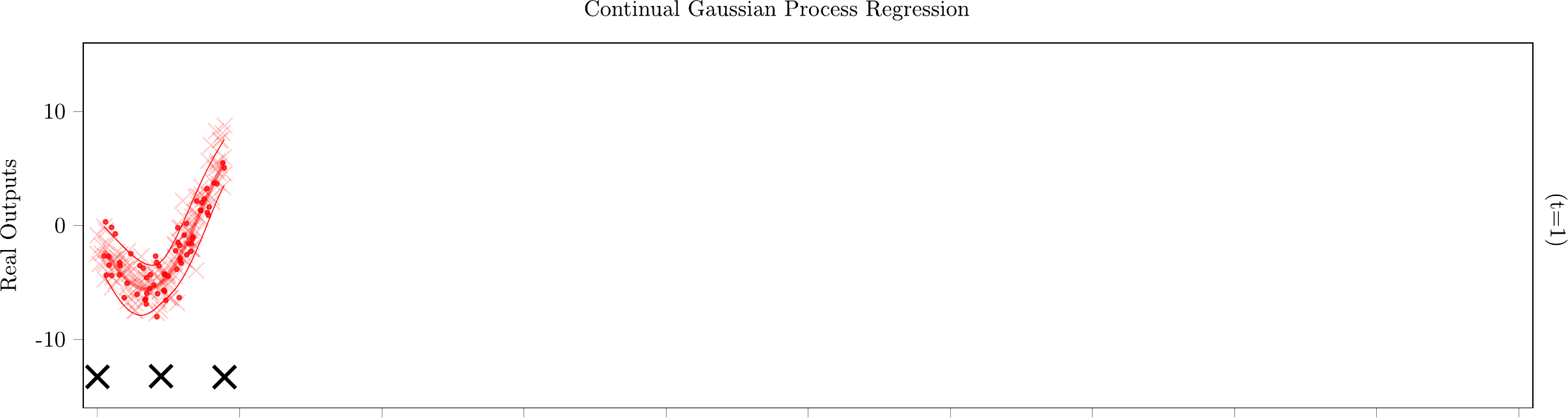}\\
		\vspace{0.1cm}
		\includegraphics[width=0.75\textwidth]{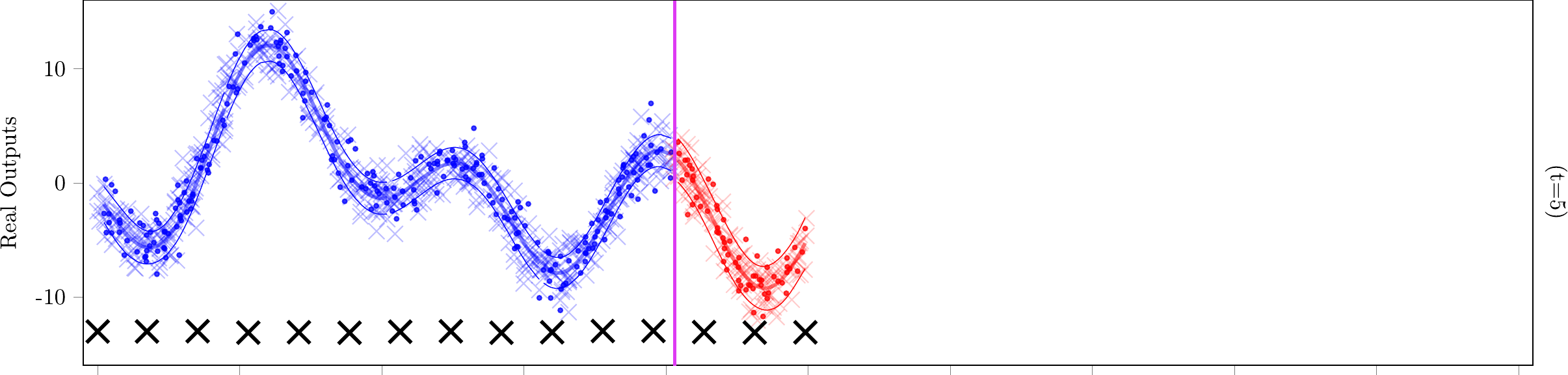}\\
		\vspace{0.1cm}
		\includegraphics[width=0.75\textwidth]{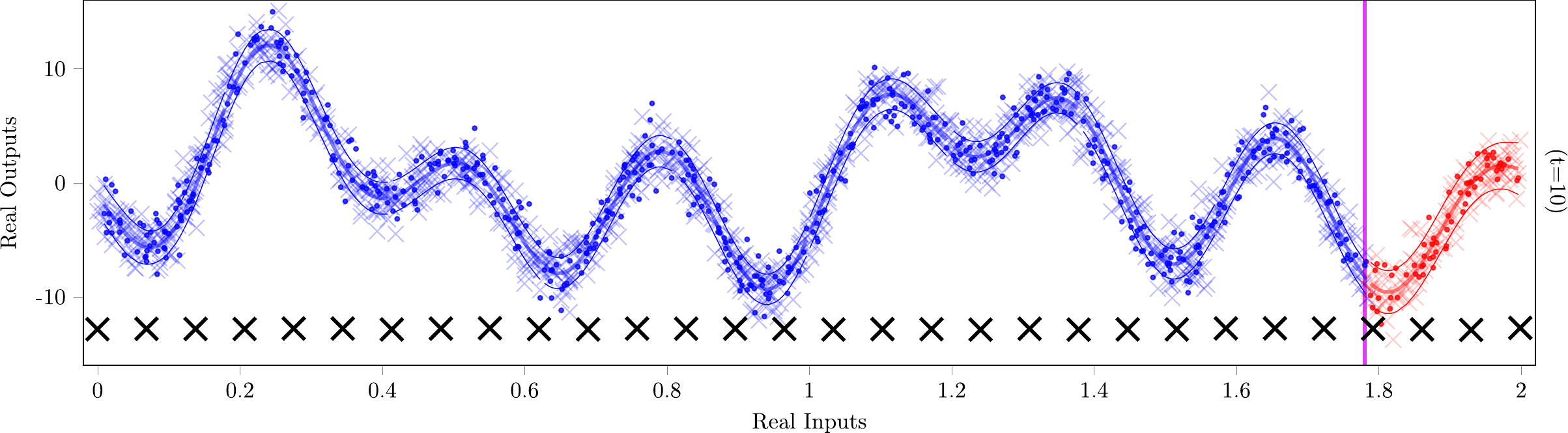}
		\caption{Results from continual GP regression applied to toy streaming data. Sequential batches correspond to non-overlapping partitions. The sequence consists of $t=10$ consecutive subsets of observations that the model acquires recursively. Red elements represent the GP predictive posterior over the newer input domain while the blue ones are refer to the past visited input space. Train and test data samples are plotted as colored crosses and dots respectively. Black crosses indicate the position of the inducing inputs at each time-step. The pink line corresponds to the limit between the past and the new input domain explored by the continual GP.}
		\label{fig:streaming}
\end{figure}

\textbf{Streaming.}~ The streaming data experiment consists of $t=10$ batches of data that are observed in a sequential manner. In this case, we consider that each batch has approximately a similar size, so the scenario is not irregular w.r.t.\ the number of samples per batch or their input domain. We setup the initial number of inducing points to be $M=3$, that will also be increased following the rule $M(t) = 3t$. The rule can be modified depending on the problem considered, as we will see later on additional experiments. We consider a synthetic dataset of $N=2000$ samples where the $30\%$ of them are used for testing. The ground-truth expression of the true latent functions is included in the Appendix. All inducing points are initialized at random in different positions based on the previous ones, that is, at time $t+1$. There are not values of $\Zcal_{\text{new}}$ that coincide with the previous ones at $\Zcal_{\text{old}}$ from the step $t$. In Figure \ref{fig:streaming}, we show three captions of the iterative learning process, concretely the initial step at $t=1$, the intermediate one at $t=5$ and the final step at $t=10$. It is important to mention that at each time-step, the posterior predictive computation of the curves does not use any past parameters, only the learned ones in the most recent iteration. Notice that, It is the last trained model, which avoids revisiting data, the one who predicts all along the input space explored so far. 

Additionally, in Table \ref{tab:streaming} we include the negative log-predictive density (NLPD) values obtained from each $t$-th subset of the test observations. All posterior predictive densities are computed via Monte-Carlo (MC) for the given selected likelihood distribution. The performance of the method is evaluated in three different ways: i) test prediction at the new observed input region, ii) decay of the predictive precision in $t$-th past seen input areas without revisiting old data samples and iii) prediction quality of the GP model all along the input domain. 

For instance, in the case of the $t'=1$ column, the NLPD is evaluated on the same test-samples as the GP model does at $t=1$. One can see how the red error metrics remain approximately static around an average NLPD value of $13.29\times 10^{-2}$ which is slightly less than the initial value obtained when data was first observed at that region. Initially, the model obtained an average of $13.13\times 10^{-2}.$ This means that, although the continual variational approach suffers a small reduction in the predictive precision once past training samples are never revisited again, the accuracy still remains constant 9 steps after its maximization, that is, 9 GP prior reconstructions and 9 optimization processes where the learned uncertainty measurements are not overwritten. One last detail is that for all metrics showed, we obtain mean and standard deviation numbers given 10 simulations with different initializations.

\begin{table}[]
	\centering
	\caption{Streaming single-output data. Test-NLPD metrics ($\times 10^{-2}$). \textbf{Column} \textsc{new}: Predictive error values obtained in the new observed input area at each time-step ($t'=t$). \textbf{Columns} \textsc{old}: Predictive error values obtained in the past observed input areas at time-steps ($t'=1, t'=4$ and $t'=8$). Colored values correspond to the GP prediction on the same test-samples at the $t$-th iteration. \textbf{Column global}: NLPD values over the test-samples all along the input domain at each time-step $t$.}
	\begin{tabular}{cccccc}
		\toprule
		& \textsc{new} & \textsc{old} & \textsc{old} & \textsc{old} &  \\
		\textbf{step} & $t'=t$ &  $t'=1$ & $t'=4$ & $t'=8$ & \textbf{global} \\
		\midrule
		$t=1$ &  \textcolor{red}{$\mathbf{13.13 \pm 0.10}$} & - & - & - & $13.13 \pm 0.13$ \\
		$t=2$ & $12.50 \pm 0.13$ & \textcolor{red}{$13.24 \pm 0.10$} & - & - & $25.74 \pm 0.23$ \\
		$t=3$ & $12.54 \pm 0.08$ & \textcolor{red}{$13.29 \pm 0.13$} & - & - & $38.48 \pm 0.27$\\
		$t=4$ & \textcolor{blue}{$\mathbf{11.59 \pm 0.04}$} & \textcolor{red}{$13.33 \pm 0.12$} & - & - & $52.26 \pm 0.28$\\
		$t=5$ & $11.34 \pm 0.05$ & \textcolor{red}{$13.28 \pm 0.10$} & \textcolor{blue}{$11.34 \pm 0.06$} & - & $63.78 \pm 0.32$\\
		$t=6$ & $11.56 \pm 0.06$ & \textcolor{red}{$13.29 \pm 0.11$} &  \textcolor{blue}{$11.33 \pm 0.06$} & - & $75.35 \pm 0.46$\\
		$t=7$ & $12.71 \pm 0.09$ & \textcolor{red}{$13.29 \pm 0.12$} &  \textcolor{blue}{$11.34 \pm 0.08$} & - & $88.09 \pm 0.55$\\
		$t=8$ & \textcolor{magenta}{$\mathbf{11.92 \pm 0.05}$} & \textcolor{red}{$13.29 \pm 0.13$} &  \textcolor{blue}{$11.34 \pm 0.06$} & - & $100.01 \pm 0.62$\\
		$t=9$ & $13.55 \pm 0.08$ & \textcolor{red}{$13.29 \pm 0.09$} &  \textcolor{blue}{$11.34 \pm 0.08$} & \textcolor{magenta}{$11.98 \pm 0.06$} & $113.60 \pm 0.58$\\
		$t=10$ & $11.73 \pm 0.06$ & \textcolor{red}{$13.30 \pm 0.14$} &  \textcolor{blue}{$11.34 \pm 0.07$} & \textcolor{magenta}{$11.97 \pm 0.04$} & $125.34 \pm 0.68$\\
		\bottomrule
	\end{tabular}
	\label{tab:streaming}	
\end{table}

\begin{figure}[] 
	\centering
	\includegraphics[width=0.75\textwidth]{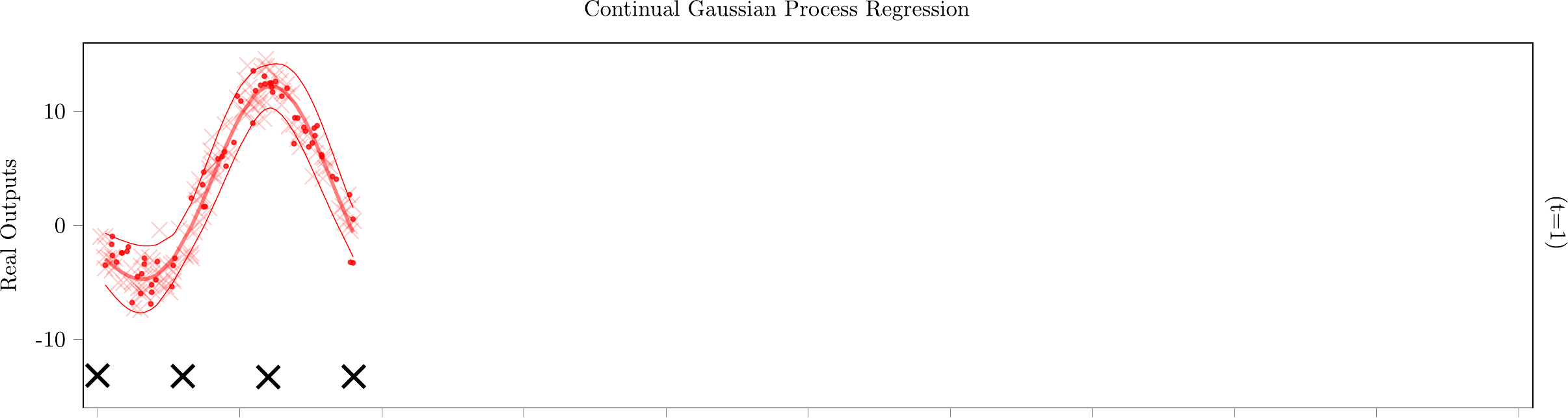}\\
	\vspace{0.1cm}
	\includegraphics[width=0.75\textwidth]{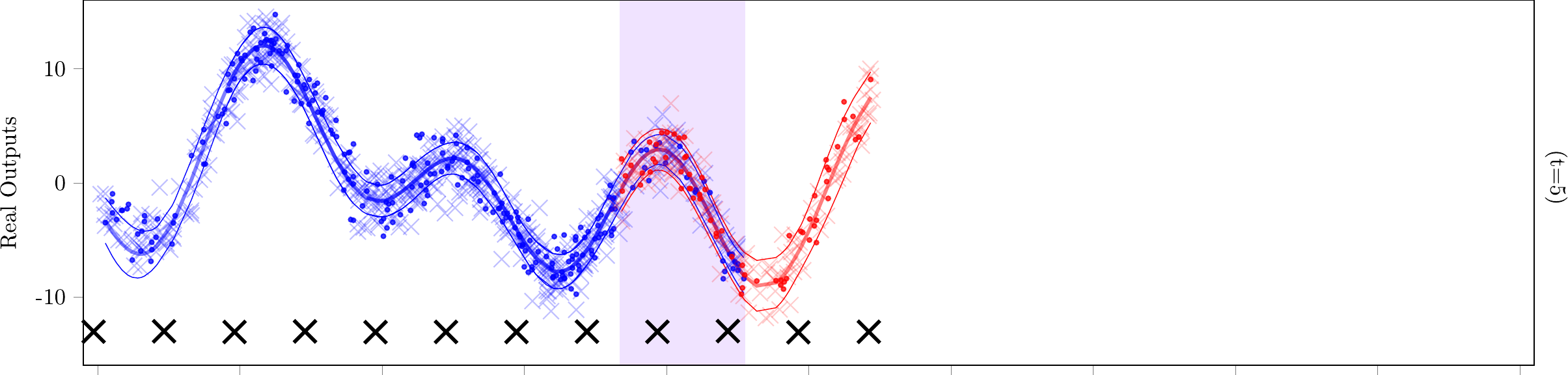}\\
	\vspace{0.1cm}
	\includegraphics[width=0.75\textwidth]{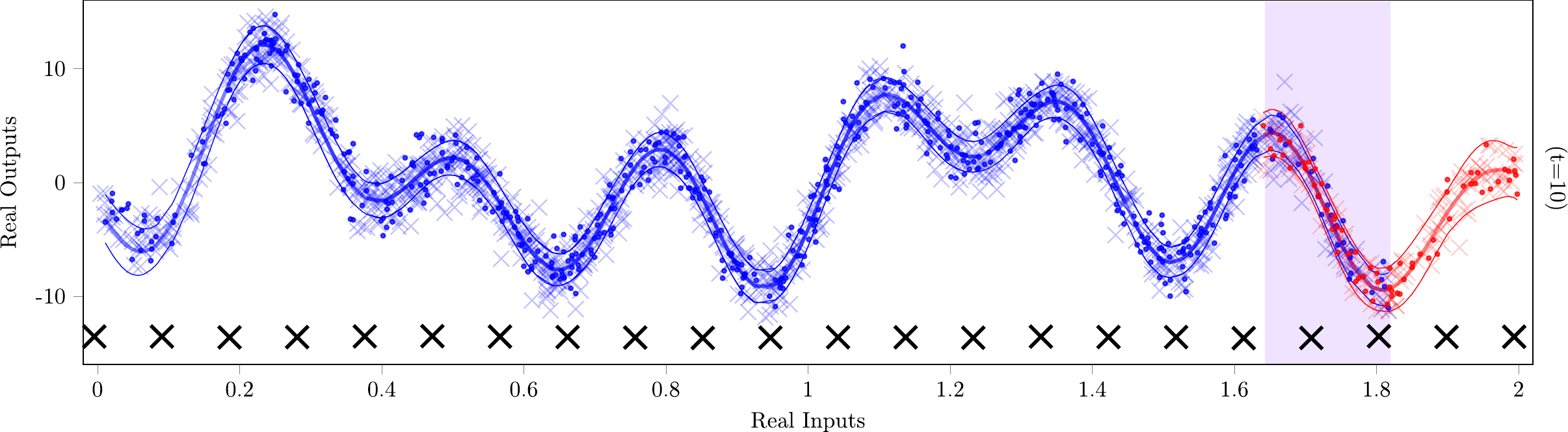}
	\caption{Three captions of the continual learning process of our single-output GP regressor. From top to down, plots correspond to steps $t=1$, $t=5$ and $t=10$. Blue and red elements correspond to past and new observed data for both training  (crosses) and test (dots) data. We consider a sequence of batches that repetitively overlaps with the last observed ones. Purple area indicates the overlapping are where past and novel data are mixed.}
	\label{fig:overlapping}
\end{figure}

\textbf{Overlapping.}~ In this version of the single-output experiment, we study the potential difficulties of the GP regression model to accept overlapping sequential batches. When we refer to overlapping partitions, we usually consider the case where a few samples revisit the input space previously observed. The setting can be observed in Figure \ref{fig:overlapping}, where we use shaded purple areas to indicate the overlapping sections of the new incoming batches. As in the previous streaming experiment, we consider a sequence of $t=10$ batches, and now the model is initialized with $M=4$ inducing points instead. The increasing rule for the sparse approximation is still linear in time steps as in the aforementioned example. Also, the learning system is limited to a maximum of 100 iterations per optimization run and importantly, the initial step of the model is trained using the standard variational bound of scalable sparse GP models \citep{hensman2015scalable,saul2016chained,morenomunoz2018}. Notice that on the first iteration, there is no past variational distribution to reconstruct the conditional GP from. 

In Table \ref{tab:overlapping}, we show similar NLPD results to the ones included in Table \ref{tab:streaming}. The first column corresponds to the NLPD metrics obtained over the new observed test-samples at the $t$-th time-step. Intermediate columns show the predictive perfomance of the GP over the past visited data. Notice that the $t'=1$ column values would correspond to the NLPD obtained by the GP at each $t$-th time-step over the input region first visited at $t=1$.

We can observe how the performance of the continual learning approach is equivalent to the streaming case. Red, blue and purple values indicate the metrics obtained once its initial training step has passed. In all cases, the precision of predictive quantities suffer an initial small reduction, but remains constant once the model continues in the number of iterations. The final number of inducing points is $M=22$.
\begin{table}[]
	\centering
	\caption{Overlapping single-output data. Test-NLPD ($\times 10^{-2}$). \textbf{Column} \textsc{new}: Predictive error values obtained in the new observed input area at each time-step ($t'=t$). \textbf{Columns} \textsc{old}: Predictive error values obtained in the past observed input areas at time-steps ($t'=1, t'=4$ and $t'=8$). Colored values correspond to the GP prediction on the same test-samples at the $t$-th iteration. \textbf{Column global}: NLPD values over the test-samples all along the input domain at each time-step $t$. In this experiment, input areas are overlapped with the previous one.}
	\begin{tabular}{cccccc}
		\toprule
		& \textsc{new} & \textsc{old} & \textsc{old} & \textsc{old} &  \\
		\textbf{step} & $t'=t$ &  $t'=1$ & $t'=4$ & $t'=8$ & \textbf{global} \\
		\midrule
		$t=1$ & \textcolor{red}{$\mathbf{13.26 \pm 0.29}$} & - & - & - & $13.26 \pm 0.29$ \\
		$t=2$ & $11.70 \pm 0.20$ & \textcolor{red}{$12.23 \pm 0.10$} & - & - & $23.94 \pm 0.30$ \\
		$t=3$ & $13.60 \pm 0.12$ & \textcolor{red}{$12.26 \pm 0.11$} & - & - & $37.58 \pm 0.31$\\
		$t=4$ & \textcolor{blue}{$\mathbf{12.63 \pm 0.13}$} & \textcolor{red}{$12.08 \pm 0.17$} & - & - & $50.37 \pm 0.50$\\
		$t=5$ & $14.50 \pm 0.36$ & \textcolor{red}{$12.07 \pm 0.12$} & \textcolor{blue}{$12.66 \pm 0.11$} & - & $64.93\pm 0.77$\\
		$t=6$ & $13.68 \pm 0.16$ & \textcolor{red}{$12.04 \pm 0.07$} & \textcolor{blue}{$12.77 \pm 0.10$} & - & $79.38 \pm 0.63$\\
		$t=7$ & $13.80 \pm 0.10$ & \textcolor{red}{$12.24 \pm 0.09$} & \textcolor{blue}{$12.75 \pm 0.12$} & - & $92.86 \pm 0.73$\\
		$t=8$ & \textcolor{magenta}{$\mathbf{13.45 \pm 0.09}$} & \textcolor{red}{$12.03 \pm 0.09$} & \textcolor{blue}{$12.67 \pm 0.11$} & - & $106.21 \pm 0.93$\\
		$t=9$ & $12.64 \pm 0.09$ & \textcolor{red}{$12.09 \pm 0.08$} & \textcolor{blue}{$12.69 \pm 0.06$} & \textcolor{magenta}{$13.78 \pm 0.09$} & $119.04 \pm 1.01$\\
		$t=10$ & $12.84 \pm 0.15$ & \textcolor{red}{$12.08 \pm 0.11$} & \textcolor{blue}{$12.71 \pm 0.08$} & \textcolor{magenta}{$13.65 \pm 0.09$} & $131.93 \pm 1.01$\\
		
		\bottomrule
	\end{tabular}
	\label{tab:overlapping}	
\end{table}

\begin{figure}[] 
	\centering
	\includegraphics[width=0.75\textwidth]{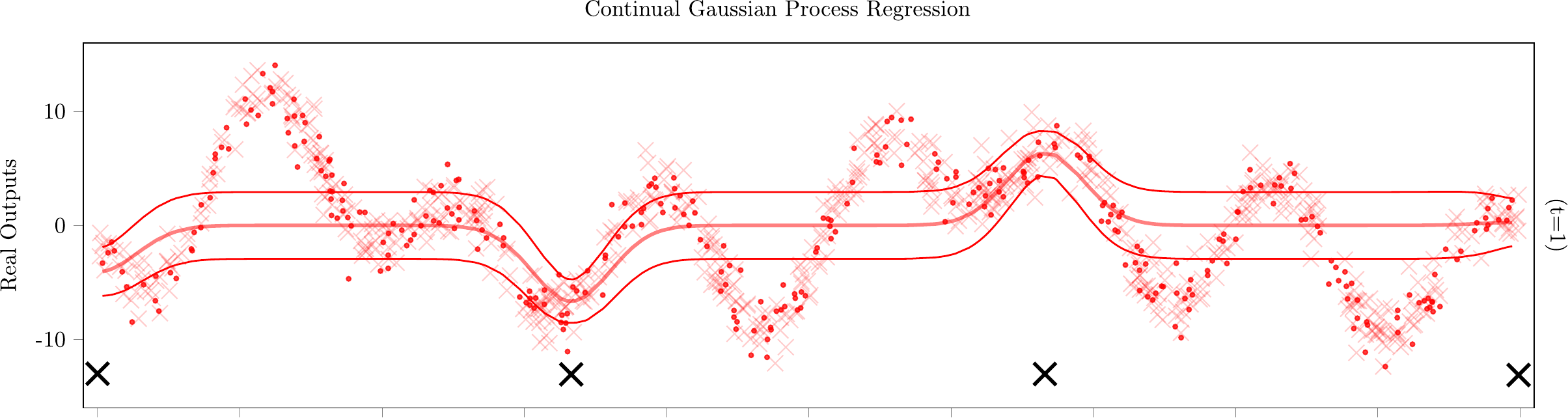}\\
	\vspace{0.1cm}
	\includegraphics[width=0.75\textwidth]{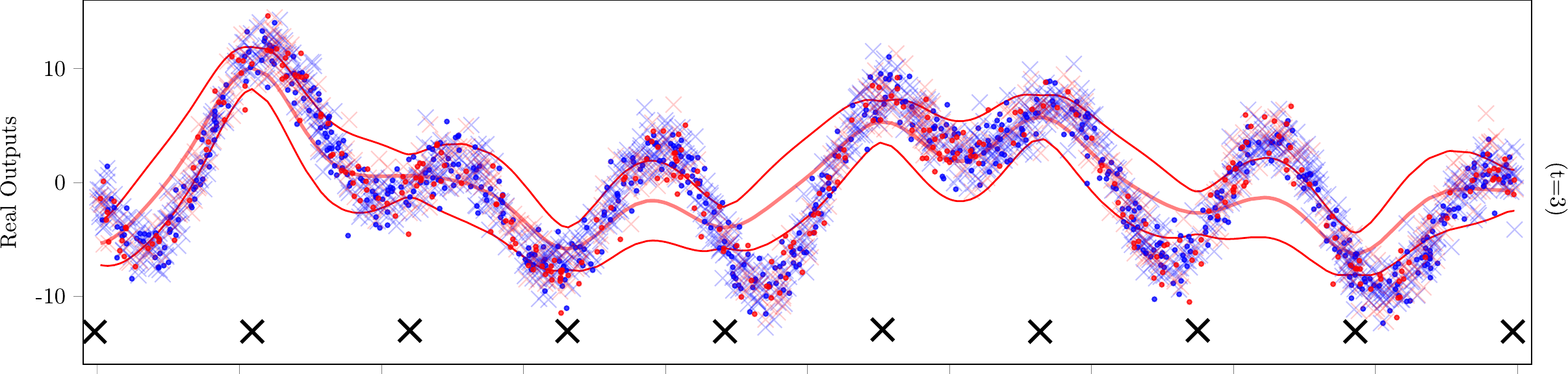}\\
	\vspace{0.1cm}
	\includegraphics[width=0.75\textwidth]{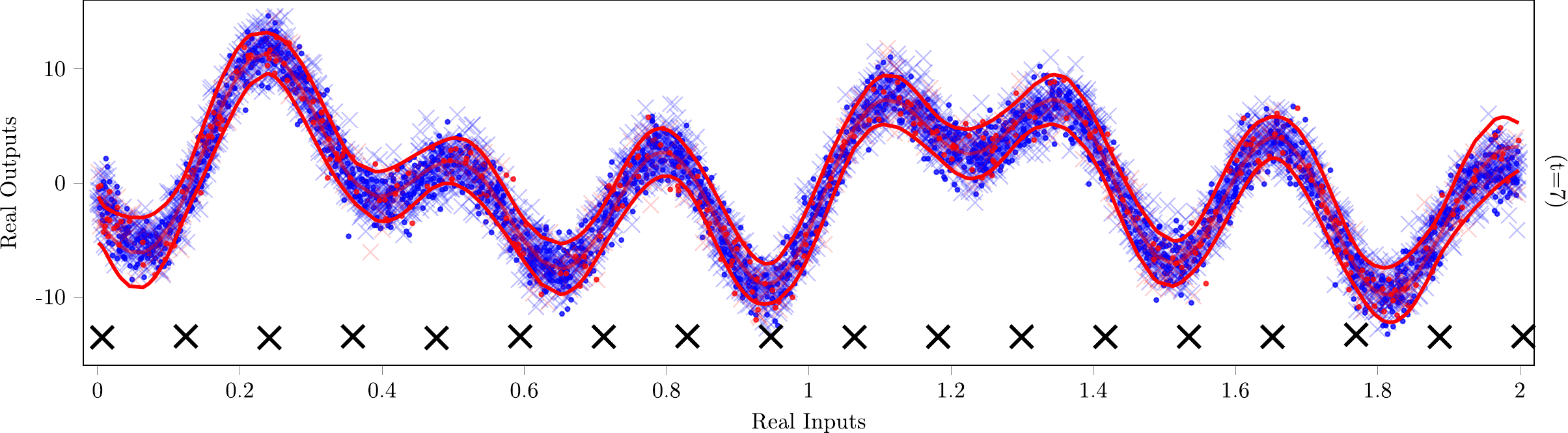}
	\caption{Representation of the continual learning process of the GP at time-steps $t=1$, $t=3$ and $t=7$. Blue and red elements correspond to past and new observed data for both training (crosses) and test (dots) data. The dataset is incrementally delivered to the learning system in small batches all along the input area. The GP model increases the number of inducing-inputs (black crosses) as long as new observations come in. Red curves indicate the posterior predictive curves over the entire input space.}
	\label{fig:incremental}
\end{figure}

\textbf{Incremental.}~ The last version of the toy single-output GP regression experiment shows relevant properties of the model itself. In this case, we setup an experiment where batches does not advance through the input space. Alternatively, we establish a pseudo-stochastic setting, where batches are observed across the entire input domain. (e.g.\ similarly to the batches used in standard SVI methods). The key point here is that we can train, reconstruct and modify the complexity of our model following any consideration observed from the new incoming data. Notice that the model allows both to increase or decrease the number of inducing points and hence, the computational cost of the variational sparse approximation. That is, in Figure \ref{fig:incremental} we can see how the number of inducing points is increased as new batches appear but exploring similar regions of the input space. At the same time, prediction curves improve as the number of inducing points increases but considering only the last observed training data so far. This is interesting for the reason that the continual mechanism is similar to SVI methods in GPs but using analytic gradients instead (use of stochastic VI implies noisy gradient vectors depending on the size of mini-batches and the learning rate hyperparameter) and it is also flexible to an irregular size of batches.

For future applications, our experiment provides a novel intuition about the potential utilities of the continual learning approach as an impreved method for stochastic approximations. Typically, when using SVI for sparse GP models, one fixes the number of inducing-inputs $M$ and applies any stochastic gradient method computed from a smaller subset of samples. However, if the sparse approximation requires a higher amount of inducing-inputs at some iteration of the learning process (e.g.\ the input domain increases), the entire GP would have to be re-defined. When using the continual GP approach, this problem disappears, as one can augment, reduce or keep constant the number $M$ of inducing-inputs. Such complexity of the sparse approximation could be chosen, for instance, using the rates in \citet{burt2019rates}. Our method also accepts SVI with an optimizer based on the stochastic gradient. In the single-output experiments, the initial number of inducing-inputs considered is $M=4$ and for this version, we set a linear rule of the form $M(t) = M(t-1) + 2t$.

In Table \ref{tab:incremental}, we show the NLPD results from the iterative process of $t=10$ steps. In contrast to the results obtained in the previous versions of the GP regression experiment, here the robustness against error propagation is not that obvious. Particularly, we can see that the prediction error values still improve after the first training iteration. This is caused by the fact that the density of inducing points is higher and also because the continual learning process is correctly propagating the posterior distribution forward. 

\begin{table}[]
	\centering
	\caption{Incremental single-output data. Test-NLPD. \textbf{Column} \textsc{new}: Predictive error values obtained in the new observed input area at each time-step ($t'=t$). \textbf{Columns} \textsc{old}: Predictive error values obtained in the past observed input areas at time-steps ($t'=1, t'=4$ and $t'=8$). Colored values correspond to the GP prediction on the same test-samples at the $t$-th iteration. \textbf{Column global}: NLPD values over the test-samples all along the input domain at each time-step $t$. In this experiment, all batches are overlapping.}
	\begin{tabular}{cccccc}
		\toprule
		& \textsc{new} & \textsc{old} & \textsc{old} & \textsc{old} &  \\
		\textbf{step} & $t'=t$ &  $t'=1$ & $t'=4$ & $t'=8$ & \textbf{global} \\
		\midrule
		$t=1$ & \textcolor{red}{$\mathbf{3.17 \pm 14.34}$} & - & - & - & $3.17 \pm 1.43$ \\
		$t=2$ & $2.58 \pm 5.69$ & \textcolor{red}{$2.56 \pm 4.71$} & - & - & $5.14 \pm 1.04$ \\
		$t=3$ & $1.46 \pm 3.70$ & \textcolor{red}{$1.22\pm 3.00$} & - & - & $3.94 \pm 1.07$\\
		$t=4$ & \textcolor{blue}{$\mathbf{1.95 \pm 6.28}$} & \textcolor{red}{$2.00 \pm 6.32$} & - & - & $7.90 \pm 2.32$\\
		$t=5$ & $1.50 \pm 2.71$ & \textcolor{red}{$1.45 \pm 3.99$} & \textcolor{blue}{$1.38 \pm 2.56$} & - & $7.16\pm 1.54$\\
		$t=6$ & $0.80 \pm 0.35$ & \textcolor{red}{$0.77 \pm 0.75$} & \textcolor{blue}{$0.79 \pm 0.85$} & - & $4.93 \pm 0.33$\\
		$t=7$ & $0.69 \pm 0.82$ & \textcolor{red}{$0.66 \pm 0.48$} & \textcolor{blue}{$0.68 \pm 0.32$} & - & $4.88 \pm 0.28$\\
		$t=8$ & \textcolor{magenta}{$\mathbf{0.63 \pm 0.23}$} & \textcolor{red}{$0.66 \pm 0.16$} & \textcolor{blue}{$0.68 \pm 0.29$} & - & $5.43 \pm 0.23$\\
		$t=9$ & $0.66 \pm 0.18$ & \textcolor{red}{$0.65 \pm 0.17$} & \textcolor{blue}{$0.66 \pm 0.18$} & \textcolor{magenta}{$0.62 \pm 0.14$} & $6.00 \pm 0.17$\\
		$t=10$ & $0.63\pm 0.16$ & \textcolor{red}{$0.64 \pm 0.13$} & \textcolor{blue}{$0.66 \pm 0.19$} & \textcolor{magenta}{$0.62 \pm 0.11$} & $6.65 \pm 0.16$\\
		\bottomrule
	\end{tabular}\\
	\footnotesize{(all std. $\times 10^{-3}$)}
	\label{tab:incremental}	
\end{table}

\textbf{Dollar Exchange Rate.}~ For our first experiment with a real-world dataset, we consider the problem of sequentially predicting a foreign exchange rate w.r.t.\ the european currency (EUR).\footnote{Currency data can be found at \url{http://fx.sauder.ubc.ca/data.html}} The setting of our experiment consists of daily ratios between the US dollar currency (USD) and Euro (EUR), taken during 48 months. The total number of samples taken is $N=922$. In this experiment, we split the dataset in 4 subsets, each subset corresponds approximately to one year. Our goal is to perform GP regression once a year without forgetting the previously learned latent functions. For the regression model, we consider a Gaussian likelihood distribution with a fixed noise parameter $\sigma=10^{-2}$ and a Mat\'ern kernel function for the GP. The applicability of the continual learning approach out of vanilla GPs. Initialization values of hyperparameters are included in the Appendix. 

\begin{figure}[] 
	\centering
	\includegraphics[width=0.8\textwidth]{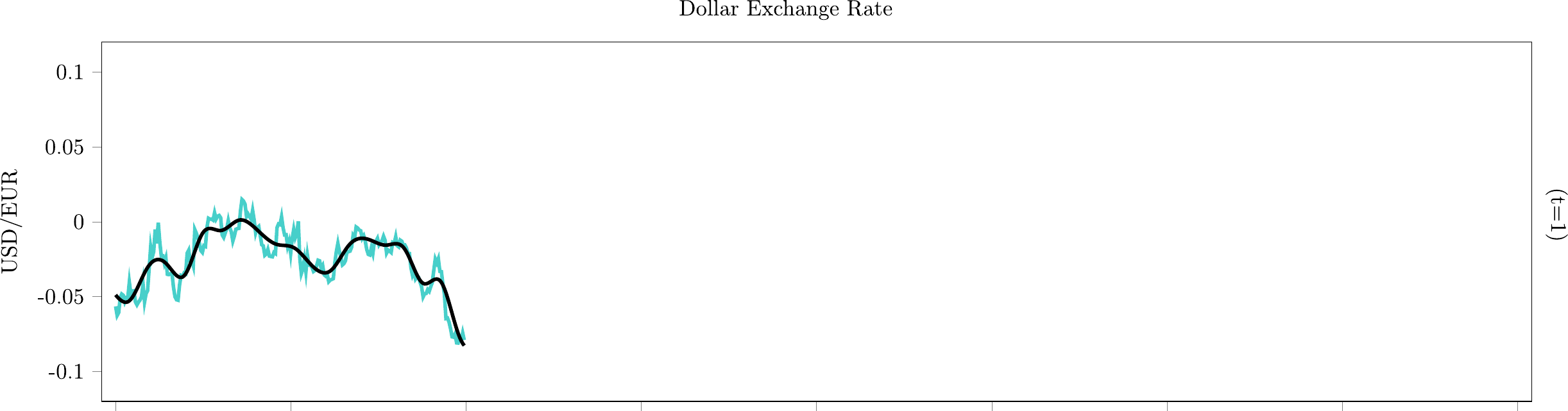}\\
	\vspace{0.1cm}
	\includegraphics[width=0.8\textwidth]{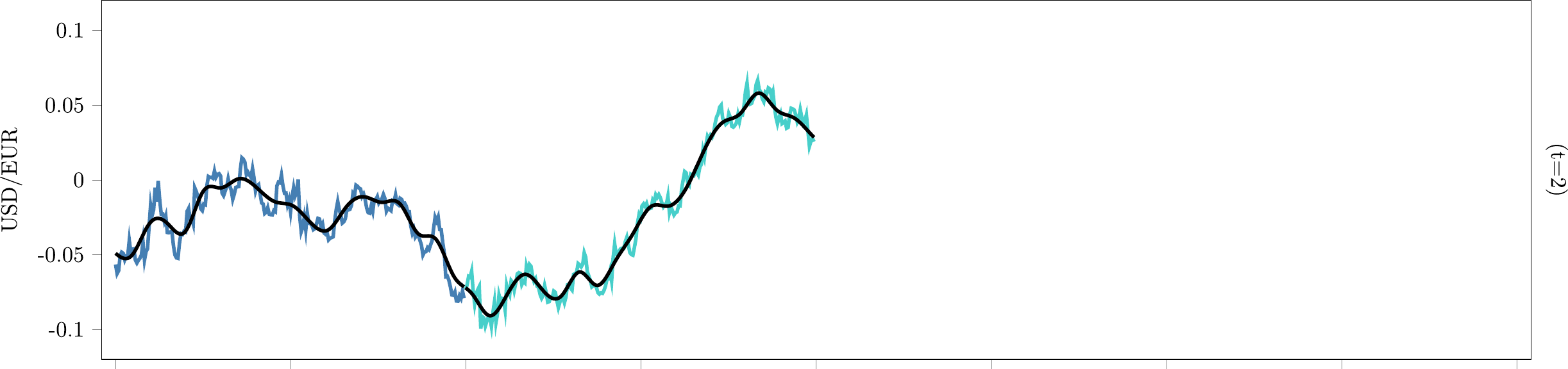}\\
	\vspace{0.1cm}
	\includegraphics[width=0.8\textwidth]{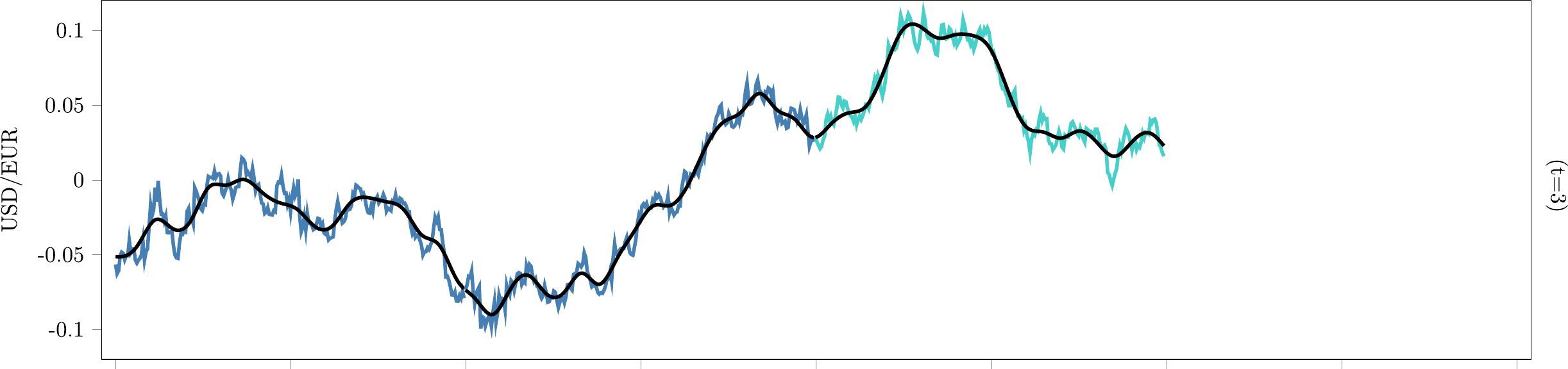}\\
	\vspace{0.1cm}
	\includegraphics[width=0.8\textwidth]{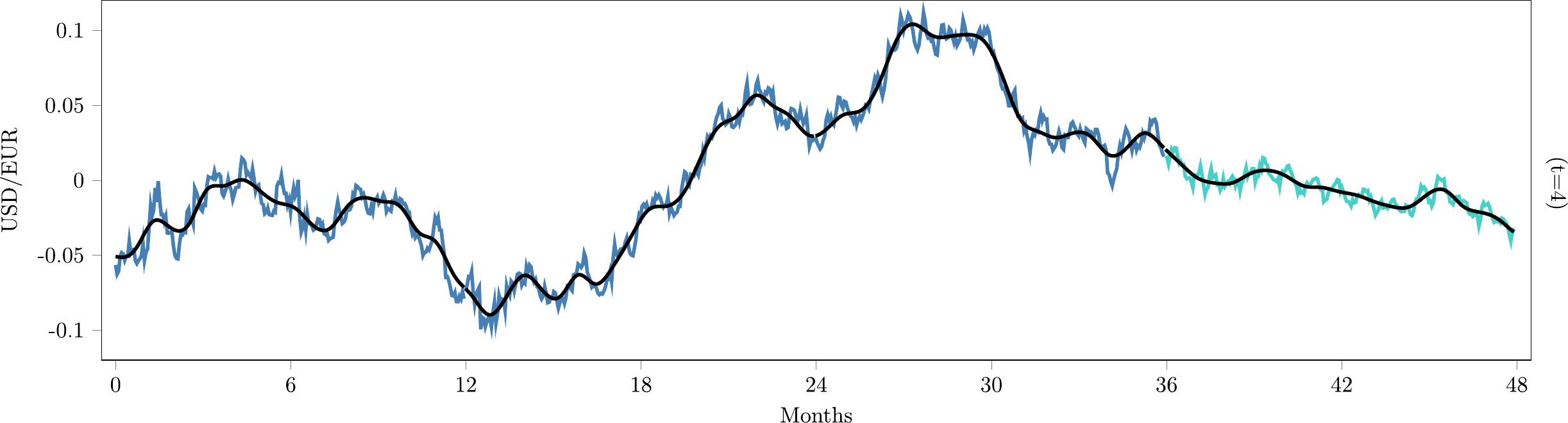}
	\caption{Evolution of the mean posterior predictive curve (black) along time under dollar exchange data. Every 12 months, the model is re-updated without revisiting past training samples. The underlying output latent function is generated from a GP prior with a Mat\'ern kernel.}
	\label{fig:currency}
\end{figure}

Similarly to Figure \ref{fig:streaming} for the toy regression experiment, in Figure \ref{fig:currency} we show 4 iterations of the sequential training process. We used different colors to indicate both old and new training samples. The GP mean predictive function (black) remains fitted all along the input domain as the model is re-trained with new data. We setup the initial number of inducing-points to $M=20$, that becomes double at each time-step.

\subsection{Continual GP classification}
\label{sec:gp_class}

The approach presented in this paper is also valid under the presence of non-Gaussian likelihood models that implies to introduce additional approximations for the computation of expectations. Hence, the expected values of likelihoods can be computed via Gaussian-Hermite quadratures if the integrals are intractable. As an example of the continual GP performance over binary data, we choose the banana dataset, used for demonstrative experiments of scalable GP classification tasks \citep{hensman2015scalable,bui2017streaming}.%

\begin{figure}[] 
	\centering
	\includegraphics[width=0.26\textwidth]{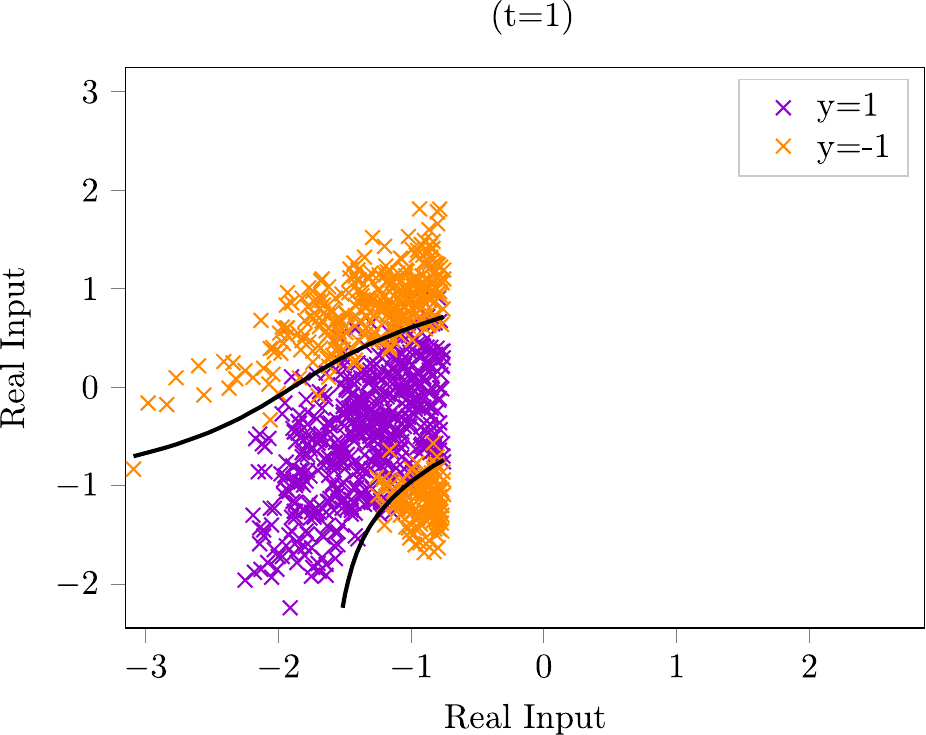}
	\includegraphics[width=0.23\textwidth]{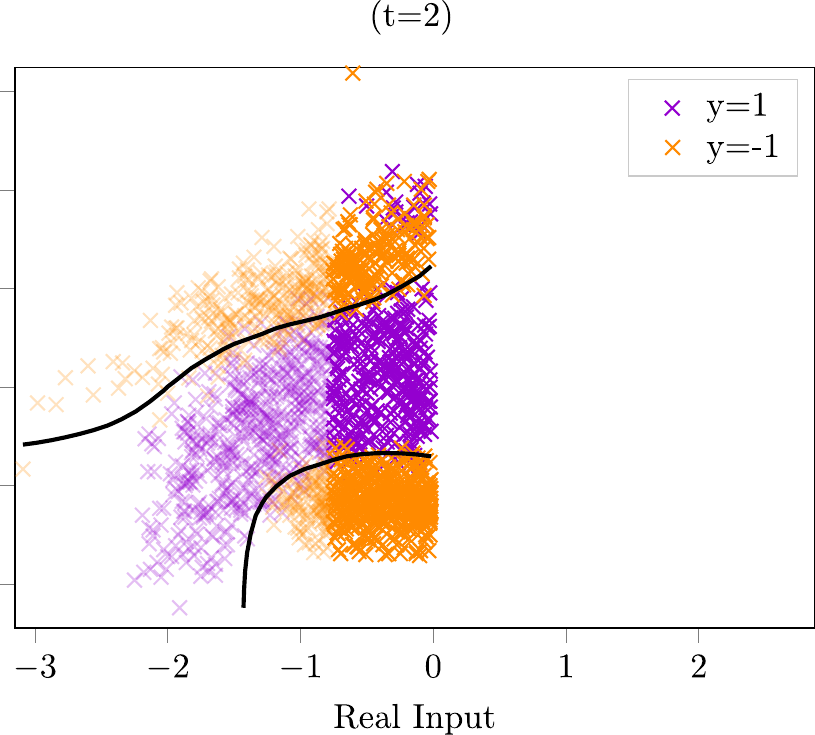}
	\includegraphics[width=0.23\textwidth]{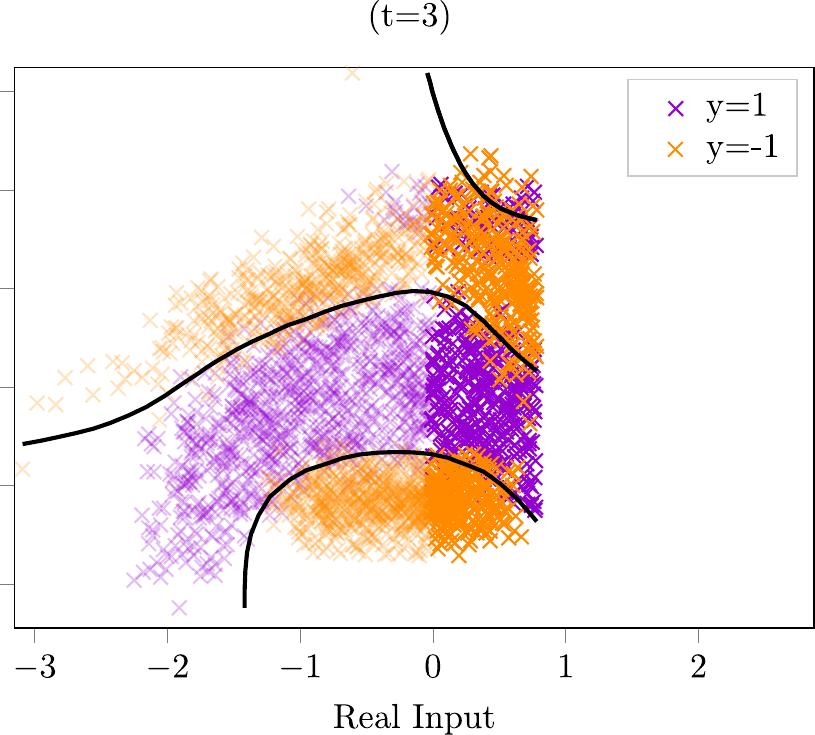}
	\includegraphics[width=0.23\textwidth]{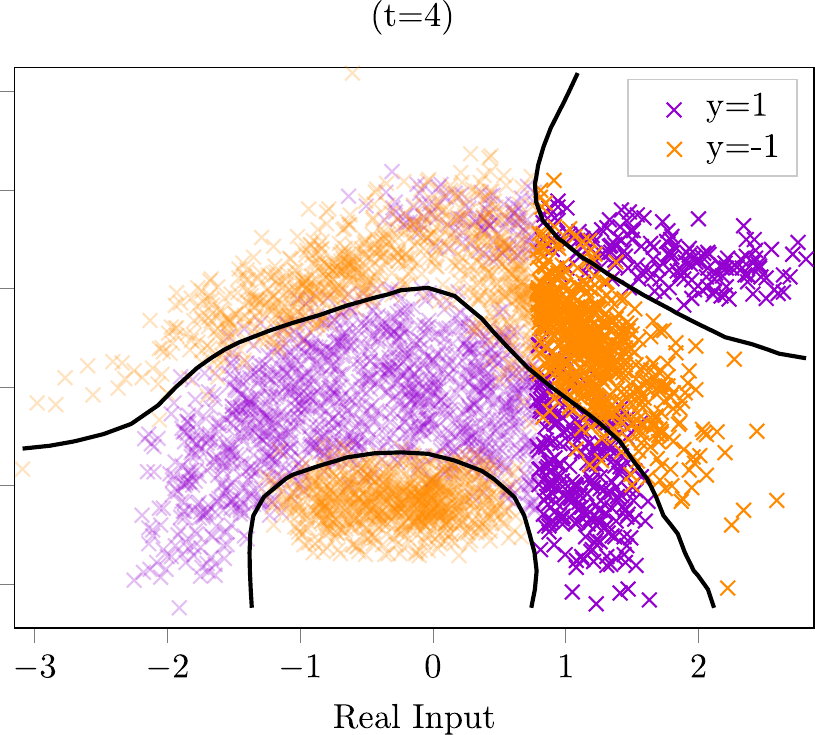}
	\caption{Performance of the continual GP learning approach under non-Gaussian data for binary classification tasks. Past samples are plotted in a grey scaled version. Black curves represent the frontier between positive and negative predictions w.r.t.\ the output values. Additionally, the last r.h.s.\ plot shows the final prediction of the model over the entire 2-dimensional input space, within the last training data seen so far (sharp colors).}
	\label{fig:banana}
\end{figure}

\textbf{Banana Dataset.}~ In the continual GP classification experiment with real-world data, we consider the case of a non-Gaussian likelihood model with an input dimensionality greater than one. Particularly, the banana dataset consists of $N=5200$ pairs of input-output observations, where we select a percentage of $30\%$ for testing the predictive error metrics. All inputs have a dimension $p=2$. In Figure \ref{fig:banana}, we plot the 4-steps inference process where we initially setup a grid of inducing points with $M=3$ inducing inputs per side. Grey scaled colors correspond to non-revisited training samples.

In Table \ref{tab:banana}, we show the NLPD results obtained in test prediction as well as the classification error rates (ER) for each time step. If we analyze the ER results, we can see that the performance is similar to the single-output GP regression case, where the precision remains constant in areas of the input space where training data is never revisited. 

\begin{table}[]
	\centering
	\caption{Banana Dataset. Test NLPD \& Classification Error Rate (ER). \textbf{Column} \textsc{new}: Predictive and error metrics obtained in the new observed input area at each time-step ($t'=t$). \textbf{Columns} \textsc{old}: Predictive and error values obtained in the past observed input areas at time-steps ($t'=1, t'=2$ and $t'=3$). Colored values correspond to the GP prediction on the same test-samples at the $t$-th iteration.}
	\begin{tabular}{cccccc}
		\toprule
		(NLPD) & \textsc{new} & \textsc{old} & \textsc{old} & \textsc{old} &  \\
		\textbf{step} & $t'=t$ &  $t'=1$ & $t'=2$ & $t'=3$ & \textbf{global} \\
		\midrule
		$t=1$ & $0.08 \pm 0.13$ & - & - & - & $0.08 \pm 0.13$ \\
		$t=2$ & $0.06 \pm 0.45$ & $0.09 \pm 7.70$ & - & - & $0.17\pm 7.20$ \\
		$t=3$ & $0.13 \pm 1.10$ & $0.09 \pm 4.90$ & $0.07 \pm 0.30$ & - & $0.30 \pm 3.40$\\
		$t=4$ & $0.09 \pm 1.10$ & $0.10 \pm 5.00$ &  $0.07 \pm 1.80$ & $0.13 \pm 1.20$ & $0.39 \pm 4.50$\\
		\midrule
		&&&&&\\
		(ER) & \textsc{new} & \textsc{old} & \textsc{old} & \textsc{old} &  \\
		\textbf{step} & $t'=t$ &  $t'=1$ & $t'=2$ & $t'=3$ &  \\
		\midrule
		$t=1$ & $0.09 \pm 0.10$ & - & - & - &  \\
		$t=2$ & $0.08 \pm 1.40$ & $0.10 \pm 9.30$ & - & - &  \\
		$t=3$ & $0.16 \pm 2.90$ & $0.10 \pm 7.80$ & $0.08 \pm 1.10$ & - & \\
		$t=4$ & $0.09 \pm 0.75$ & $0.10 \pm 12.4$ &  $0.08 \pm 2.10$ & $0.14 \pm 3.80$ & \\
		\bottomrule
	\end{tabular}\\
	\footnotesize{all std. ($\times 10^{-3}$)}
	\label{tab:banana}	
\end{table}

\subsection{Continual multi-output GPs} 

As we explained in Section \ref{sec:continual_mogp}, the multi-output framework introduces two layers in the inference mechanism. One is related to the latent functions $\Ucal$, where the sparse approximation lies, while the other comes from the observational side, where expectations are evaluated from output functions $\Fcal$. The two layers make the continual multi-output learning process work in a different manner w.r.t.\ the marginal lower bound $\Lcal_{\mathcal{C}}$. Now, the expectation terms are \textit{decoupled} from the regularization side which is only focused on the latent function priors. The key property of the continual multi-output approach is that we can consider extremely irregular problems where, for instance, outputs are completely asymmetric as we will show in the following results. An illustration of the asymmetric cases can be seen in Figure \ref{fig:mo_cases}. In this section of experiments, we include three cases, two of them using toy regression data and a third one with real-world observations from human motion capture.

\begin{figure}[] 
	\centering
	\includegraphics[width=0.75\textwidth]{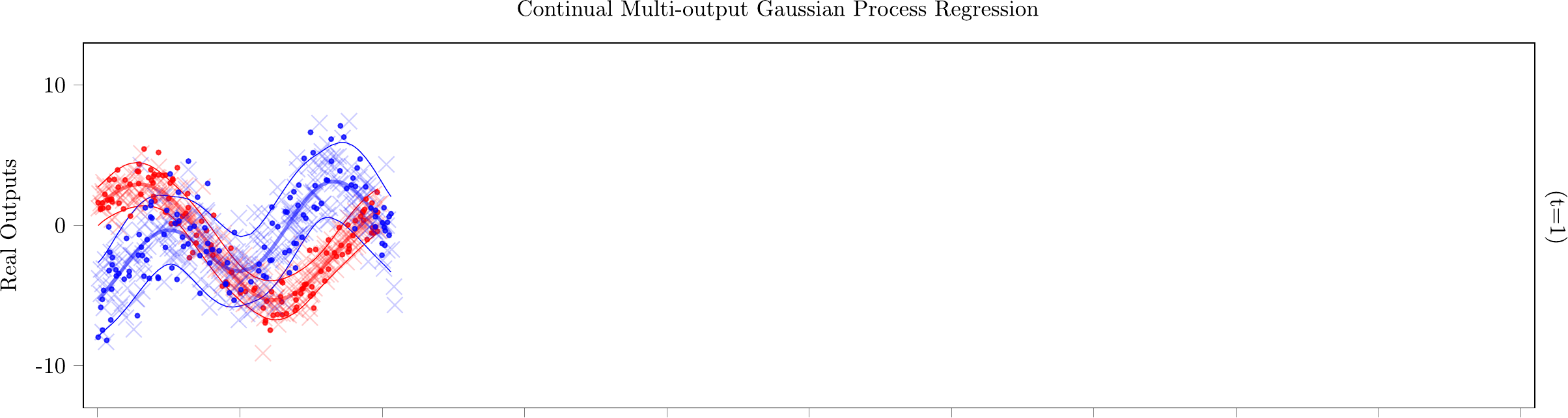}\\
	\vspace{0.1cm}
	\includegraphics[width=0.75\textwidth]{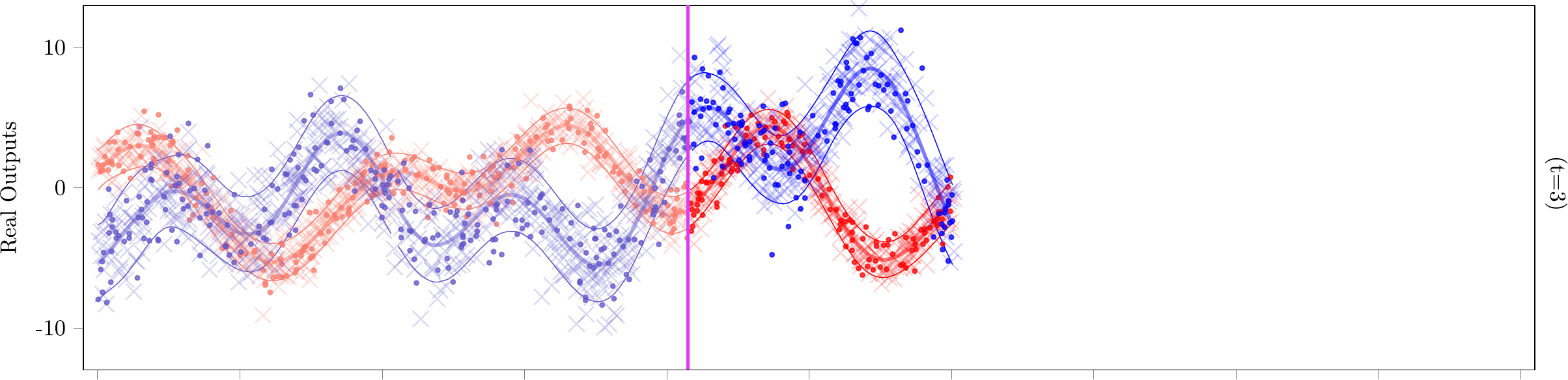}\\
	\vspace{0.1cm}
	\includegraphics[width=0.75\textwidth]{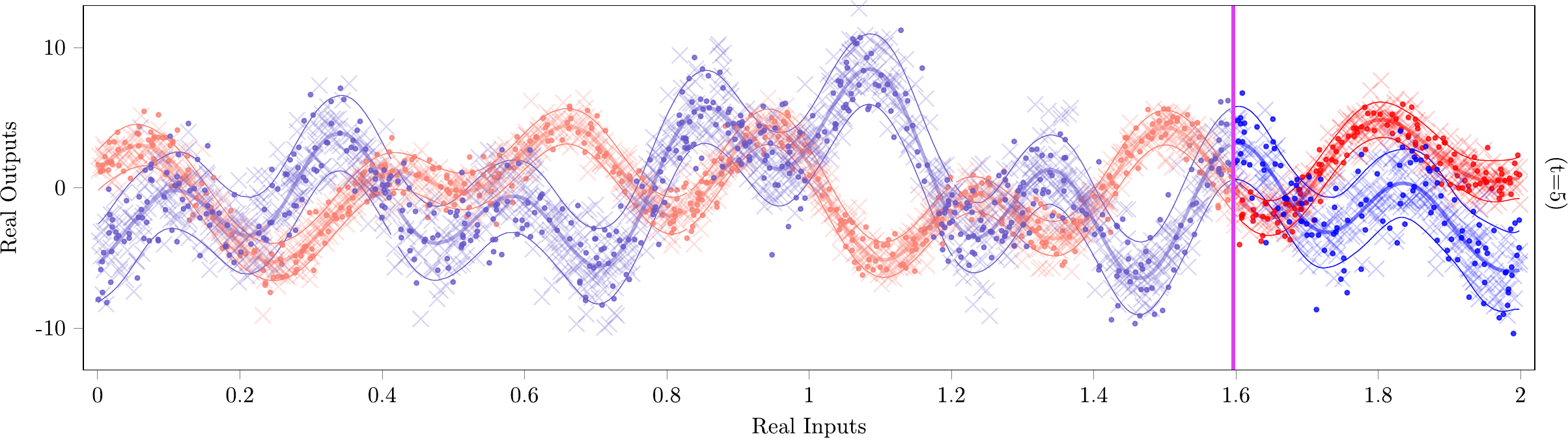}
	\caption{Results for temporal modeling of multi-output real-valued data. Two channels are jointly model using the continual learning approach aforementioned for multi-output GP regression. The pink line indicates the limiting point between the novel observed samples and the past data that we avoid to revisit. All inducing-inputs are positioned over the $Q$ underlying latent functions that are later combined to obtain the output parameter functions. Both channels are trained together in a synchronous manner. The $Q$ subsets of inducing-inputs are not plotted for a reason of clarity.}
	\label{fig:synchronous}
\end{figure}

\textbf{Synchronous Channels.}~ In the first multi-output experiment with toy data, we are interested into jointly performing multi-task non-linear regression over two output Gaussian channels with different likelihood noise parameters. The underlying linear mixing of the latent functions is assumed to follow a LMC structure that we also aim to infer it in an online manner. The number of true latent functions is $Q=2$ and we generate them using a linear combination of sinusoidal signals (see details in Appendix). In this case, we have artificially split the dataset into five batches of non-overlapping samples that are delivered sequentially at the same time-step on both channels. In Figure \ref{fig:synchronous}, we show three captures of the learning process for this experiment. Additionally, the empirical error results for test prediction are included in Table \ref{tab:synchronous}, where the predictive error metrics are equivalent to the ones obtained in the previous single-output cases. 

\begin{table}[]
	\centering
	\caption{Synchronous multi-channel streaming data. Test-NLPD (all std. $\times 10^{-4}$). \textbf{Columns} \textsc{new}: Predictive error values obtained in the new observed input area at each time-step ($t'=t$) for each channel. \textbf{Columns} \textsc{old}: Predictive error values obtained in the past observed input areas at time-step $t'=1$ for both channels. Colored values correspond to the GP prediction on the same test-samples at the $t$-th iteration. \textbf{Columns global}: NLPD values over the test-samples all along the input domain at each time-step $t$ and channel.}
	\begin{tabular}{ccccccc}
		\toprule
		channel $\rightarrow$& I &II& I& II & I & II\\
		& \textsc{new} & \textsc{new}  & \textsc{old} & \textsc{old} &  &\\
		\textbf{step} & $t'=t$ &  $t'=t$ & $t'=1$ & $t'=1$ & \textbf{global} & \textbf{global}\\
		\midrule
		$t=1$ & \textcolor{red}{$\mathbf{0.19 \pm 0.36}$} & \textcolor{blue}{$\mathbf{0.30 \pm 2.81}$} & - & - & $0.19 \pm 0.07$ & $0.30 \pm 0.56$ \\
		$t=2$ & $0.18 \pm 0.53$ & $0.35 \pm 2.07$ & \textcolor{red}{$0.19 \pm 0.71$} & \textcolor{blue}{$0.32 \pm 2.53$} &$0.38 \pm 0.25$& $0.67 \pm 0.92$ \\
		$t=3$ & $0.19 \pm 0.42$ & $0.40\pm 1.64$ & \textcolor{red}{$0.19 \pm 0.48$} & \textcolor{blue}{$0.31 \pm 1.97$} & $0.58 \pm 0.27$& $1.07 \pm 1.13$\\
		$t=4$ & $0.17 \pm 0.49$ & $0.33 \pm 1.66$ & \textcolor{red}{$0.19 \pm 0.83$} & \textcolor{blue}{$0.31 \pm 1.98$} &$0.75 \pm 0.45$ & $1.41 \pm 1.58$\\
		$t=5$ & $0.16 \pm 0.37$ & $0.35 \pm 1.81$ & \textcolor{red}{$0.19 \pm 0.29$} & \textcolor{blue}{$0.31 \pm 2.19$} &$0.92 \pm 0.38$ &$1.76\pm 1.93$\\
		\bottomrule
	\end{tabular}\\
	\footnotesize{($^*$) colors correspond to output channels in Figure \ref{fig:synchronous}}. 
	\label{tab:synchronous}	
\end{table}

\textbf{Asynchronous Channels.}~ The following experiment is of particular importance for the demonstration of the multi-output model performance under asymmetric incoming channels. Particularly, we consider the same dataset as in the synchronous scenario but introducing an asymmetric observation process over the incoming channels data by the learning system. That is, at each time-step, only one of the two channels delivers output-input samples. In the next step, the observation channel switches and new incoming data appears on the other one. This observation procedure is depicted in Figure \ref{fig:asynchronous}.

The continual inference process is possible due to the latent functions $\Ucal$ lie in a different layer than the output observations. Hence, the inducing points can be positioned across the input domain within the emergence of new samples in any of the output channels. The number of initial inducing points is $M_q = 4$ per channel, and double per time-step iteration.

\begin{figure}[] 
	\label{fig:asynchronous}
	\centering
	\includegraphics[width=1.0\textwidth]{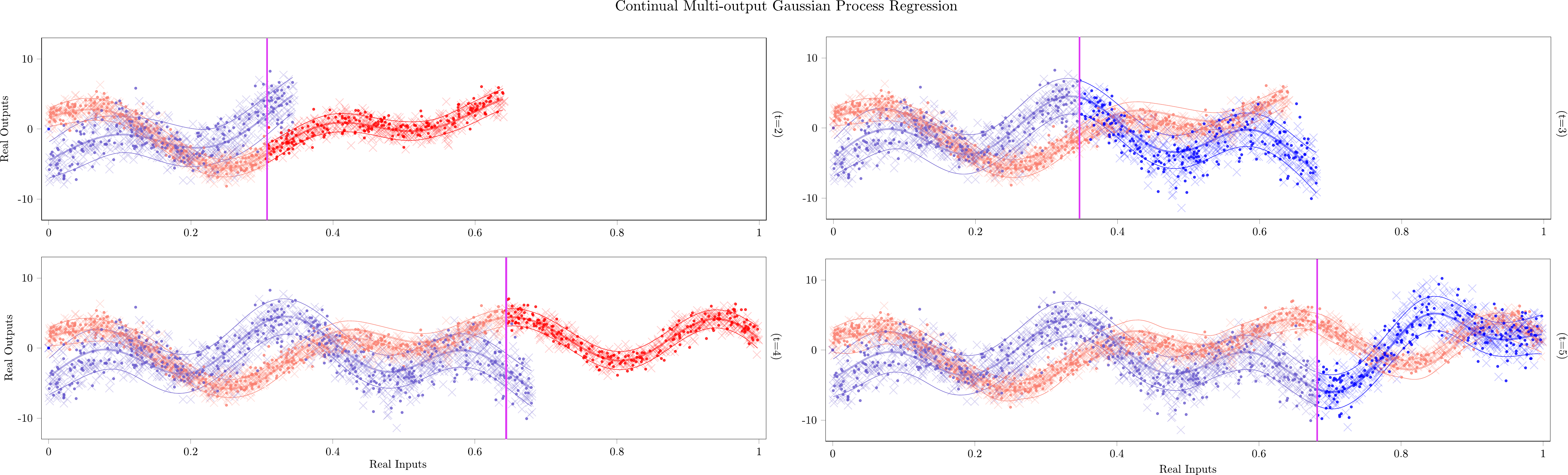}
	\caption{In contrast to Figure \ref{fig:synchronous}, we apply the continual GP approach to model multi-channel sequential data that is observed in an asynchronous manner, that is, samples might appear at different time steps from different outputs in unobserved input regions. From left to right and from top to down, we represent the learning process at four consecutive time-steps ($t=2$, $t=3$, $t=4$ and $t=5$). Past data is plotted using grey scaled colors.}
	\label{fig:asynchronous}
\end{figure}

\textbf{Multi-channel sensors for Human Motion.}~ For the last multi-output regression experiment with real-world data, we consider the MOCAP dataset.\footnote{MOCAP datasets are available at \url{http://mocap.cs.cmu.edu/.}} The data consists of raw multi-channel traces from sensors monitoring human motion. In particular, we select the first individual (id. number $01$) in the \textit{walking} activity example. We aim to exploit the benefits of multi-task GPs rather that using a single-output GP per sensor. It is demonstrated that by exploiting such correlations between channels, multiple-output data are better modelled \citep{bonilla2008multi}. From all available sensors in the human body, we consider three of them whose oscillation phase does not coincide: the \textit{left wrist}, the \textit{right wrist} and at the \textit{right femur}. Each channel provides a number of $N=343$ samples corresponding to the vertical axis values recorded by the sensors. For the experiment, we setup an initial amount of $M=10$ inducing inputs in order to obtain a reliable precision. We increase the $M$ twice per recursive iteration. Moreover, the number of latent functions in the multi-output GP prior is $Q=3$. Both latent function values and the underlying linear mixing coefficients are initialized at random at each time-step.

\begin{figure}[] 
	\centering
	\includegraphics[width=0.75\textwidth]{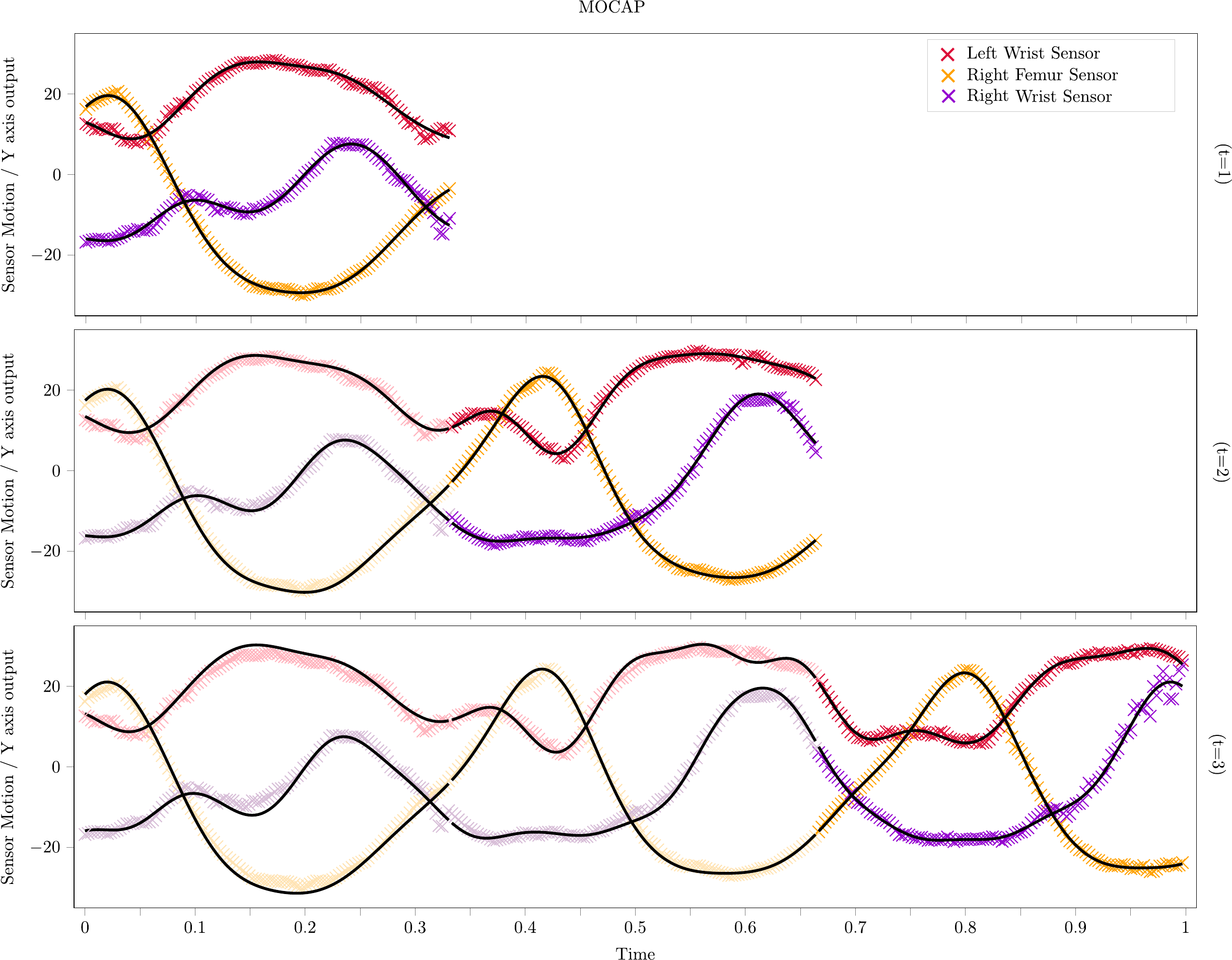}
	\caption{\textsc{MOCAP} dataset. Multi-output GP regression over three sequential channels. Each channel corresponds to the Y axis output values of a sensor in a walking motion capture experiment. Black curves correspond to the mean of the posterior predictive distribution at each time-step for the whole input space. Gray scaled colors correspond to non-revisited data samples.}
	\label{fig:mocap}
\end{figure}

The multi-output model with the LMC formulation is robust. It recovers the previous linear combination from random initial values thanks to the triple KL regularization within the continual MOGP prior. In Figure \ref{fig:mocap} we show the performance of the multi-task regression model for the three regression outputs at 3 different time-steps. Each color represents a different sensor channel. 

\subsection{Resistance to propagation error}

In this experiment, we are particularly interested in the demonstration of the effect that the continual GP prior reconstruction has on the whole model. In particular, how robust it can be as $t\rightarrow \infty$. Typically, substituting variational posterior distributions $q(\cdot)$ as the novel prior into a Bayesian online updating scheme seems the most natural manner to treat sequential observations using approximated probabilistic inference. However, this approach is usually discarded due to the assumption that repeated approximations may accumulate errors as the number of time-steps increases \citep{nguyen2017variational}, something that usually happens. 

One of the main objectives in our work is to beat this assumption, performing continual variational learning for signal processing applications with thousands of updating repetitions. In the following experiment, we present some results that aim to demonstrate this statement. We also prove that recursively reconstructing the continual GP prior avoids propagating the error of approximations forwards.

\textbf{Solar Physics Data.}~ Based on filtering experiments for signal processing applications, we obtained an astrophysics dataset which consists of the monthly average of sunspot counting numbers from 1700 to 1995. In particular, we use the observations made for the analysis of sunspot cycles by the Royal Greenwich Observatory (US).\footnote{Solar physics data is publicly available at \url{https://solarscience.msfc.nasa.gov/}} For avoiding the use of non-tractable likelihood models, we transform the strictly positive samples into the real domain by means of the non-linear mapping $\log(1+\xc)$. Note that the original observations are the average of counting numbers obtained from several observers. 

\begin{figure}[] 
	\centering
	\includegraphics[width=0.75\textwidth]{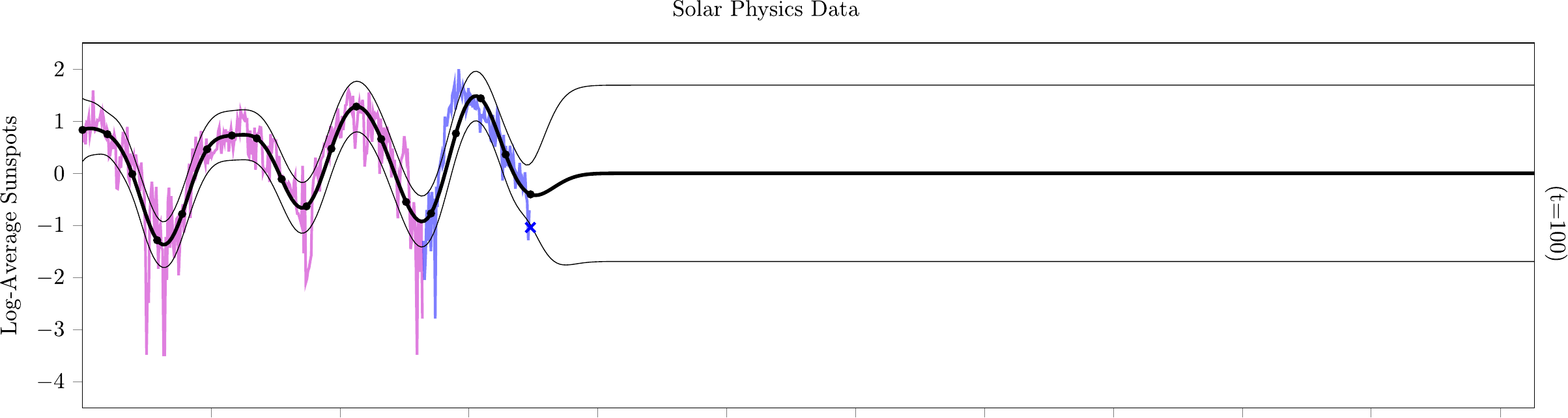}\\
	\vspace{0.1cm}
	\includegraphics[width=0.75\textwidth]{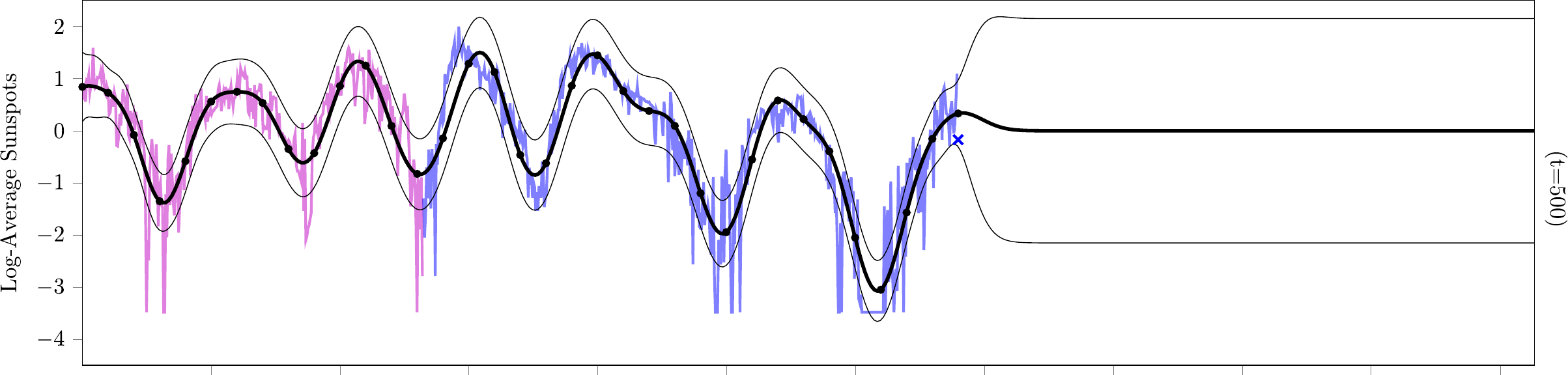}\\
	\vspace{0.1cm}
	\includegraphics[width=0.75\textwidth]{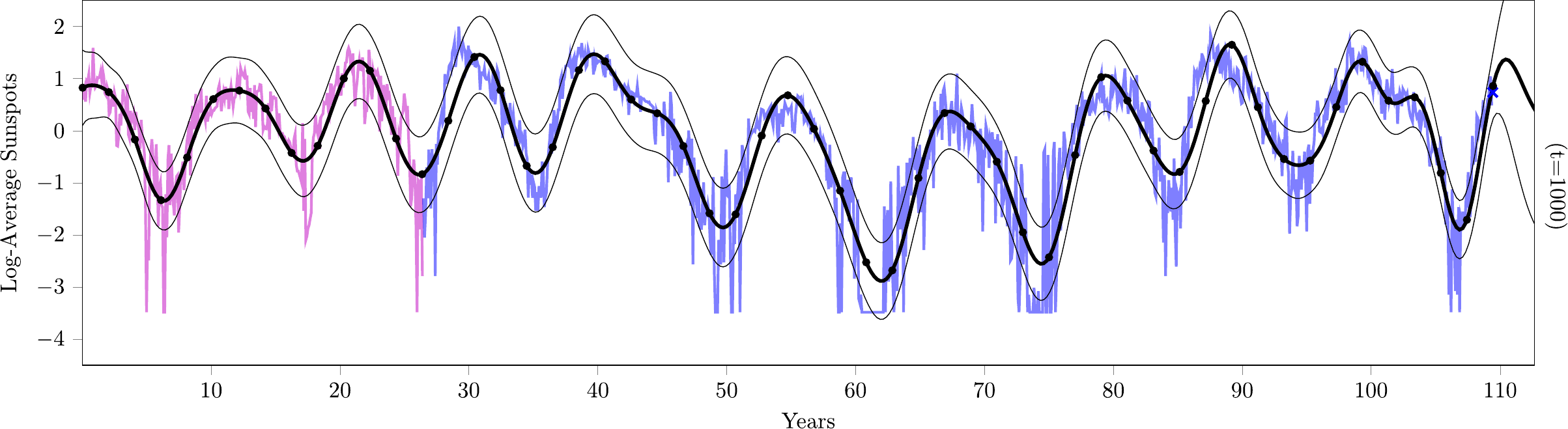}
	\caption{Results for single-output regression on solar physics data with one-sample updates of the continual sparse GP model. Pink colored signal corresponds to the \textit{warm up} observations in the batch mode. Greyed blue signals correspond to the former visited observations while the blue cross is the new incoming one. Black colored curves correspond to the mean function and the 95\% confidence interval of the predictive GP distribution all over the input-space, computed at each time iteration. Black dots are the inducing variables at each time-step.}
	\label{fig:solar}
\end{figure}

Our primary goal is to demonstrate that the predictive mechanism of the continual GP remains stable when $t\rightarrow \infty$, all over the input domain, i.e.\ it does not forget past visited regions. In Figure \ref{fig:solar}, we show three captures of the continual learning process until a maximum of $t=10^3$ iterations. It is important to mention that we used a one-sample update rule for the entire sequence, meaning $10^3$ consecutive optimization trials. For tractable reasons, we setup an initial number of $M=10$ inducing points for the \textit{warm up} period and an incremental update of one additive inducing point per 100 new samples observed. We also included a similar transition for the parameters and initialization points as in the previous experiments.

A demonstrative visualization of the whole continual GP learning process for the solar sunspot signal can be found at \url{https://www.youtube.com/watch?v=j7kpru4YrcQ}. Importantly, the predictive GP posterior distribution remains accurate and fitted to the signal without revisiting data during $t=10^3$ iterations. 

\subsection{Continual GP vs. Baseline methods}

In our last experiment, we are interested in the comparison of the continual GP framework with previous baselines techniques in the literature. As we mentioned in our revision of the state-of-the-art, the works that our approach is most related to are: i) the infinite-horizon Gaussian process (IHGP) in \citet{solin2018infinite} and ii) the streaming sparse Gaussian process (SSGP) in \citet{bui2017streaming} for the single-output case. %

\begin{figure}[]
	\centering
	\includegraphics[width=\textwidth]{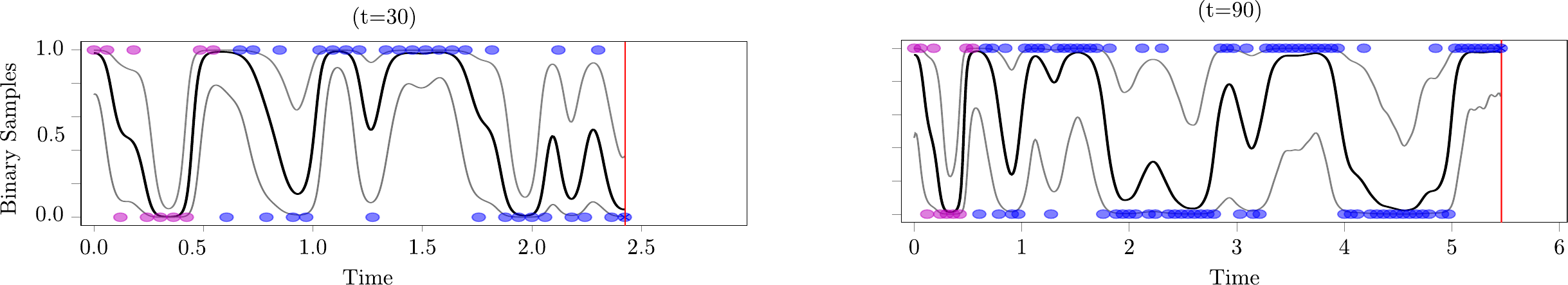}
	\caption{Results for continual single-output GP classification over probit toy data \citep{solin2018infinite}.}
	\label{fig:horizon}
\end{figure}

\textbf{Infinite-Horizon Gaussian Processes.} We test the continual GP model under the same toy experiment included in \citet{solin2018infinite} for GP classification. The initial hyperparameters are set equal to the IHGP. An important difference w.r.t.\ the aforementioned baseline model is that the IHGP focuses exclusively on accurate online predictions \textit{forward} rather than the \textit{backward} memory of the model for the already seen input-domain. For that reason, we aim to demonstrate that the continual GP approach is able to predict in an \textit{online} classification task similarly as the IHGP model does. In Figure \ref{fig:horizon}, we show the results for $t=30$ and $t=90$ in a total of 100 time-steps. The fitting accuracy is similar to the one showed by the IHGP model. Importantly, we recursively perform one-sample updates of the model, to adapt the continual GP for a most similar scenario to the one presented in the IHGP toy experiment.

\begin{figure}[]
	\centering
	\includegraphics[width=\textwidth]{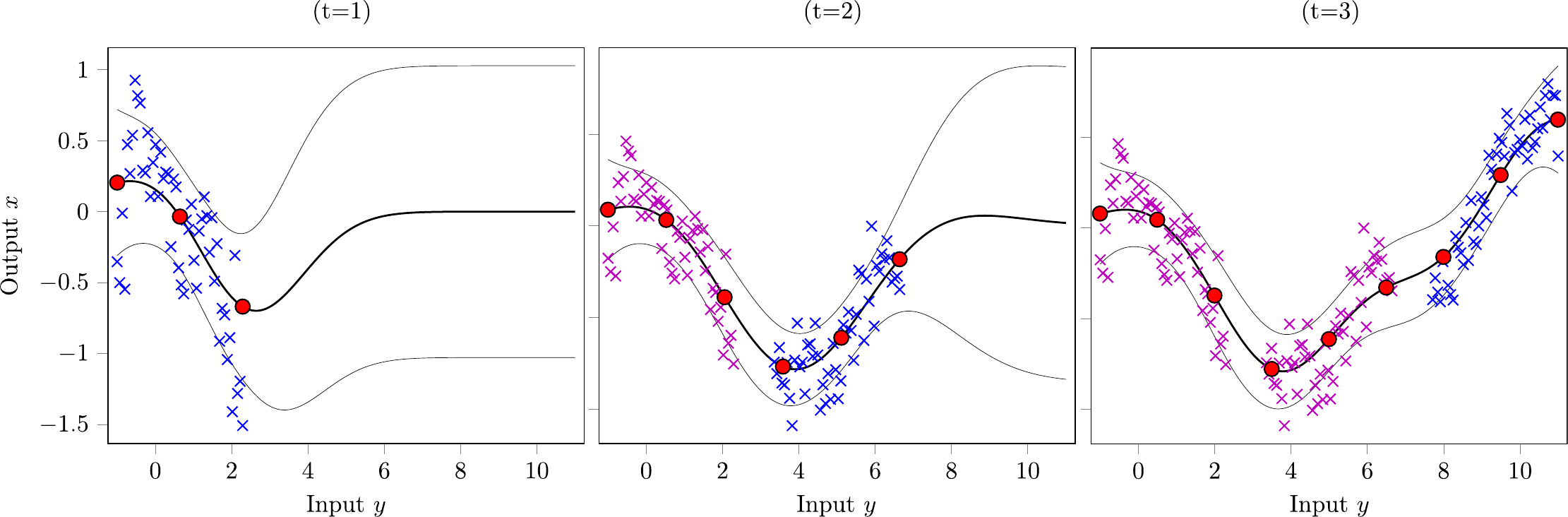}
	\caption{Results for continual single-output GP regression over real-valued toy data \citep{bui2017streaming}. Magenta and blue crosses correspond to past and new observed output samples, respectively. Red bullets are the inducing variables $\unew$ at each time-step ($t=1$, $t=2$ and $t=3$).}
	\label{fig:streaming_bui}
\end{figure}

\textbf{Streaming Sparse Gaussian Processes.} For the second comparative experiment, we test our continual GP on the two datasets used in \citet{bui2017streaming}. The first one is the banana dataset for sparse GP classification. The results and classification error metrics are included in the experiment of Section \ref{sec:gp_class} and Figure \ref{fig:banana}. In the second case, we take the toy regression data from its Github code. \footnote{Toy data available at \url{https://github.com/thangbui/streaming_sparse_gp}.} We imitate the setup of the SSGP toy experiment where the sequence of observations is split in three partitions, with $M=3$ inducing points per partition. In Figure \ref{fig:streaming_bui}, we show three captures of the results for the predictive curves of the GP regression model. We also plot the position of the inducing points (red bullets) as a proof that the continual GP method is analogous to SSGP when applied under the same scenario. The only existing difference is that our single-output model recursively builds the continual GP prior instead of concatenating old and new inducing-points $\u$, that tends to be less robust as the input domain augments.

\section{Conclusion and Future Work}

\textbf{Conclusion.}~ In this paper, we have presented a novel approach that extends the existing posterior-prior recursion of online Bayesian inference to the infinite functional framework of Gaussian process models. The key principle of our continual learning method is that we are able to reconstruct implicit GP priors over the space-of-functions conditioned to past posterior distributions via the predictive GP formulation. We adapt the entire method for accepting sparse approximations based on inducing-inputs for a reason of scalability. The recursive inference mechanism makes possible to update global posterior distributions without the necessity of unfeasible training computations or data revisiting. Thus, we only require to propagate the past learned parameters forward, rather than concatenating old and new data for avoiding model forgetting. Moreover, our method is fully scalable and amenable for stochastic variational inference both on regression and classification problems with arbitrary likelihood functions. Another point of interest is its simplicity when applied to the multi-output GP setting. In this case, we have shown the main differences with the single-output model, and its applicability to scenarios with asymmetric channels or even heterogeneous likelihoods, that is, mixed classification and regression problems.

\textbf{Contribution.}~ The main novelty of our work is on the recursive construction of the GP prior conditioned to the fitted variational posterior distribution. The idea of building continual GP priors, instead of concatenating inducing-points in a sparse approximation context had not been considered before. Similar uses of the predictive formula within the posterior distribution were analyzed in \citet{girard2003gaussian} before the appearance of variational methods in the GP literature. The recursive construction of GPs is equivalent to the posterior-prior recursion of online Bayesian inference. Additionally, the chance of handling a new continual GP prior makes the current approach feasible to multi-output scenarios where otherwise, concatenating inducing points would not be possible. 
 
\textbf{Future work.}~ We find that our continual learning scheme has important connections with other recent works in variational inference methods. For instance, with \citet{ruiz2019contrastive} and their contrastive divergence (VCD) based on three KL divergence terms. The idea of a triple regularized bound also emerges naturally in our continual learning problem from the Bayes rule when avoiding data revisiting. It can be easily interpreted as the difference between two divergences that balance contributions of some variational posterior distribution w.r.t.\ different objectives. However, as \citet{ruiz2019contrastive} explains, the subtraction of two KL divergences might not satisfy the properties of a divergence operator (to be always non-negative and becoming zero if equal), something that breaks the consistency of the bound and \textit{a priori} is problematic. Fortunately, adding an extra \textit{force} to the subtraction of divergences, that is, the third KL term between both variational distributions, reduces the discrepancy and makes the operator consistent for the log-marginal lower bound in a similar way to our solution. 

Future research lines are, for instance, to employ convolutional processes (CPs) or non-linear mappings as the mixing operator in the multi-output GP model as an alternative to the LMC. Moreover, the continual single-output GP model could be used as a latent baseline in the multivariate time series imputation method of \citet{fortuin2019multivariate}, which uses a GP to capture temporal dependencies between real-valued latent variables that are later connected to a deep sequential variational autoencoder (VAE). Another promising work would be to study the need of  increasing the number $M$ of inducing points as the input domain augments. It could be specified via the recent bounds for sparse approximations proposed in \citet{burt2019rates}. Finally, we may adapt both the single- and the multi-output continual model to accept non-stationary latent functions similarly to \citet{zhang2019sequential} or even infinite number of latent GP functions via mixture of experts \citep{pradier2018infinite}.

\subsection*{Acknowledgements}

PMM acknowledges the support of his FPI grant BES-2016-077626 from the Ministerio of Econom\'ia of Spain. AAR was supported by the Ministerio de Ciencia, Innovaci{\'o}n y Universidades under grant TEC2017-92552-EXP (aMBITION), by the Ministerio de Ciencia, Innovaci{\'o}n y Universidades, jointly with the European Commission (ERDF), under grant  RTI2018-099655-B-I00 (CLARA), and by The Comunidad de Madrid under grant Y2018/TCS-4705 (PRACTICO-CM). MAA has been financed by the EPSRC Research Projects EP/R034303/1 and EP/T00343X/1.


\newpage

\appendix
\section*{Appendix A. Complete derivation of continual lower bounds}
\label{app:derivation}

\textbf{Single-output GP.}~ To derive the continual lower bound for each iteration of the sequential process, we use the following expression

\begin{eqnarray}
\log p(\yc) &=& \log \int p(\yc|f)p(f)df =  \log \int p(\ycnew, \ycold|f)p(f)df  \\
&=&  \log \int p(\ycnew| f) p(\ycold|f) p(f)df \geq \mathcal{L}_{\mathcal{C}} \\
\mathcal{L}_{\mathcal{C}} &=&  \int \log p(\ycnew| f) p(\ycold|f) p(f)df = \int q(f|\phinew) \log \frac{p(\ycnew| f) p(\ycold|f) p(f)}{q(f|\phinew)}df \\
&=& \int q(f|\phinew) \log \frac{p(\ycnew| f) q(f|\phiold) p(f|\psinew)}{p(f|\psiold)q(f|\phinew)}df \\
&=& \int q(f|\phinew) \log \frac{p(\ycnew| f) p(f_{\neq u_{*}}|u_{*}, \psiold)\widetilde{q}(u_{*}|\phiold) p(f|\psinew)}{p(f|\psiold)q(f|\phinew)}df \\
&=& \int q(f|\phinew) \log \frac{p(\ycnew| f) p(f_{\neq u_{*}}|u_{*}, \psiold)\widetilde{q}(u_{*}|\phiold) p(f_{\neq u_{*}}|u_{*}, \psinew) p(u_{*}|\psinew)}{p(f_{\neq u_{*}}|u_{*}, \psiold)p(u_{*}|\psiold)p(f_{\neq u_{*}}|u_{*}, \psinew)q(u_{*}|\phinew)}df \\
&=& \int q(f|\phinew) \log \frac{p(\ycnew| f) \widetilde{q}(u_{*}|\phiold)  p(u_{*}|\psinew)}{ p(u_{*}|\psiold)q(u_{*}|\phinew)}df \\
&=& \int q(f|\phinew) \log p(\ycnew| f)df - \int q(f|\phinew) \log \frac{q(u_{*}|\phinew)}{p(u_{*}|\psinew)}df +\int q(f|\phinew) \log \frac{\widetilde{q}(u_{*}|\phiold)}{p(u_{*}|\psiold)}df \nonumber \\
&=& \int q(f_{\neq \{\fnew, u{*}\}}, \fnew, u_{*} |\phinew) \log p(\ycnew| \fnew)f_{\neq \{\fnew, u{*}\}} d\fnew du_{*} \nonumber \\
&-& \int q(f_{\neq u{*}}, u_{*}|\phinew) \log \frac{q(u_{*}|\phinew)}{p(u_{*}|\psinew)} df_{\neq u_{*}} du_{*} + \int q(f_{\neq u{*}}, u_{*}|\phinew) \log \frac{\widetilde{q}(u_{*}|\phiold)}{p(u_{*}|\psiold)} df_{\neq u_{*}} du_{*} \\
&=& \int q(u_{*}|\phinew)p(\fnew|u_{*}) \log p(\ycnew| \fnew )d\fnew du_{*} - \int q(u_{*}|\phinew) \log \frac{q(u_{*}|\phinew)}{p(u_{*}|\psinew)}du_{*} \nonumber\\
&+&\int q(u_{*}|\phinew) \log \frac{q(u_{*}|\phinew) \widetilde{q}(u_{*}|\phiold)}{q(u_{*}|\phinew) p(u_{*}|\psiold)}du_{*},
\end{eqnarray}
where we assume $u_{*}$ to be the new subset of inducing-points $\unew$, then
\begin{eqnarray}
&=& \int q(\fnew) \log p(\ycnew| \fnew)d\fnew - \int q(\unew|\phinew) \log \frac{q(\unew|\phinew)}{p(\unew|\psinew)}d\unew \nonumber\\
&+& \int q(\unew|\phinew) \log \frac{q(\unew|\phinew)}{p(\unew|\psiold)}d\unew - \int q(\unew|\phinew) \log \frac{q(\unew|\phinew)}{\widetilde{q}(\unew|\phiold)}d\unew \\
&=& \mathbb{E}_{q(\fnew)}[\log p(\ycnew|\fnew)] - \text{KL}[q(\unew|\phinew) || p(\unew|\psinew) ] + \text{KL}[q(\unew|\phinew) || p(\unew|\psiold) ] \nonumber\\
&-& \text{KL}[q(\unew|\phinew)||\widetilde{q}(\unew|\phinew) ].
\end{eqnarray}

It is important to rely on the variational expectation terms for the likelihood where $q(\fnew)$ intervenes. Particularly, we can take explicit vector values $\unew$ for the implicit inducing points notation $u_{*}$. The general expectation integral takes the form

\begin{eqnarray}
	\int q(u_{*}|\phinew)p(\fnew|u_{*}) \log p(\ycnew| \fnew )d\fnew du_{*} 
	&=& \int q(\u|\phinew)p(\fnew|\unew) \log p(\ycnew| \fnew )d\fnew d\unew \nonumber \\
	&=& \int q(\u|\phinew)p(\fnew|\unew) d\unew \log p(\ycnew| \fnew )d\fnew \nonumber \\
	&=& \int q(\fnew) \log p(\ycnew| \fnew )d\fnew,
\end{eqnarray}
and considering we denote $q(\fnew)$ as the expected variational distribution over the output vector $\fnew$, that can be analytically calculated as follows
\begin{eqnarray}
	q(\fnew) &=& \int q(\unew|\phinew)p(\fnew|\unew) d\unew \nonumber\\
	&=& \Ncal (\fnew| \K_{\fnew\unew}\K^{-1}_{\unew\unew}\bm{\mu}_{\text{new}}, \K_{\fnew\fnew} + \K_{\fnew\unew}\K^{-1}_{\unew\unew}(\S_\text{new} - \K_{\unew\unew})\K^{-1}_{\unew\unew}\K^{\top}_{\fnew\unew} ). \nonumber
\end{eqnarray}


\section*{Appendix B. Continual GP priors}
\label{app:gradients}

\textbf{Single-output GP.}~ To sequentially evaluate the approximated lower bound on our marginal likelihood distribution, we have to reconstruct the continual prior using the conditional predictive formula of GP models. Assuming that $q(\uold|\phiold)$ is our past learned variational distribution and we want to infer the probability values on an implicit vector $u_{*}$ of inducing points; the continual GP prior follows the expression
\begin{eqnarray}
	\widetilde{q}(u_{*}|\phiold) &\approx& \int p(u_{*}|\uold)q(\uold|\phiold)d\uold \nonumber\\
	&=& \Ncal (u_{*}| k_{*\uold}\K^{-1}_{\uold\uold}\bm{\mu}_{\text{old}}, k_{**} + k_{*\uold}\K^{-1}_{\uold\uold}(\S_\text{old} - \K_{\uold\uold})\K^{-1}_{\uold\uold}k^{\top}_{*\uold}), \nonumber \\
\end{eqnarray}
\noindent and if we assume that $u_{*} = \unew$, this is, evaluate the conditional predictive distribution on the future inducing points $\unew$, the previous formula takes the form of a Gaussian distribution whose expression is
\begin{equation}
	\widetilde{q}(\unew|\phiold) = \Ncal (\unew | \K_{\unew\uold}\K^{-1}_{\uold\uold}\bm{\mu}_{\text{old}}, \K_{\unew\unew}  + \K_{\unew\uold}\K^{-1}_{\uold\uold}(\S_\text{old} - \K_{\uold\uold})\K^{-1}_{\uold\uold}\K^{\top}_{\unew\uold}). \nonumber
\end{equation}

\textbf{Multi-output GP.}~ For the multiple output case, the derivation of the continual GP expression is analogous but considering the two-layers scheme. This means that the continual mechanism of reconstruction now works directly on the $Q$ underlying latent functions $u_q$, that are modeled independently. Therefore, the closed-form distribution can be obtained as
\begin{equation}
\widetilde{q}(\u_{q,\text{new}}|\phiold) = \Ncal (\unew | \K_{\unew\uold}\K^{-1}_{\uold\uold}\bm{\mu}_{\text{old}}, \K_{\unew\unew}  + \K_{\unew\uold}\K^{-1}_{\uold\uold}(\S_\text{old} - \K_{\uold\uold})\K^{-1}_{\uold\uold}\K^{\top}_{\unew\uold}). \nonumber
\end{equation}

\section*{Appendix C. Dimensionality reduction of $p(f)$ via Gaussian marginals.}

We use the properties of Gaussian marginals to reduce infinite dimensional distributions $p(f)$. This process is applied for both GP priors $p(f)$ and the Gaussian variational distribution $q(f)$. We assume that if the generative process of latent functions is $f\sim p(f)$, then it also holds

\begin{equation*}
\begin{bmatrix}
f_{\neq\unew}\\\unew
\end{bmatrix} \sim p(f_{\neq\unew}, \unew),\\
\end{equation*}
where the multivariate Gaussian distribution $p(f_{\neq\unew}, \unew)$ has the following $\K$ and $\bm{\mu}$ parameters
\begin{equation*}
p(f_{\neq\unew}, \unew) = \mathcal{N}\Big(\begin{bmatrix}
\bm{\mu}_{f\neq\unew}\\\bm{\mu}_\unew
\end{bmatrix}, \begin{bmatrix}
\K_{f_{\neq\unew}f_{\neq\unew}} \; \K_{f_{\neq\unew}\unew}\\\K_{\unew f_{\neq\unew}} \; \K_{\unew \unew}
\end{bmatrix}\Big),\\
\end{equation*}
and we therefore, may apply the marginalization $p(\unew)$ to obtain the target Gaussian distribution
\begin{equation*}
\int p(f_{\neq\unew}, \unew) df_{\neq\unew} = p(\unew) = \mathcal{N}(\bm{\mu}_\unew, \K_{\unew \unew}).
\end{equation*}


\section*{Appendix D. Experiments and hyperparameter setup}
\label{app:gradients}

The code for the experiments is written in Python and publicly available. It can be found in the repository \url{https://github.com/pmorenoz/ContinualGP}, where we extend the \texttt{HetMOGP} tool from \citet{morenomunoz2018} to be applied over sequences of multiple-output observations. Importantly, all NLPD metrics in Section \ref{sec:experiments} are computed from a total of $10^3$ samples in 10 different initializations. To make our experiments fully reproducible, we provide the details for all the experiments as well as the initializing values for all parameters and hyperparameters.

\textbf{Streaming.}~ We use a sequence of $N=2000$ toy observations, that is split into $T=10$ batches. The train-test data rate is $33\%$ for the test samples. The initial number of inducing-points $M=3$ and we use the rule $M_t = tM$ at each time-step. Additionally, we use an RBF kernel function $k(\cdot,\cdot)$ whose hyperparameters, i.e.\ length-scale and amplitude are always initialized at $\ell=0.01$ and $\sigma_a=0.5$. We assume that the likelihood function is a Gaussian distribution with a fixed noise parameter $\sigma_n=1.5$. Additionally, the true underlying functions $f$ is generated by mixing three sinusoidal signals, its expression is\\
\begin{equation*}
	f(x) = \frac{9}{2}\cos(2\pi x + \frac{3\pi}{2}) - 3\sin(4.3\pi x + \frac{3\pi}{10}) + 5\cos(7\pi x + 2.4\pi).
\end{equation*}

\textbf{Overlapping.}~ The setup of the second toy single-output experiment is analogous to the previous one but with a few exceptions. The initial number of inducing points is $M=4$, and we increase its capacity by setting $M_t = 2tM$. The kernel function and the initialization of parameters is equal to the \textit{streaming} experiment. The overlapping sections are generated by randomly indexing observations from the adjacent partitions.\\

\textbf{Incremental.}~ The setup of the \textit{incremental} experiment is analogous to the previous ones. In this case, we randomly index observations to generate the sequence of batches. The initial number of inducing-points is $M=4$ and increases similarly to the \textit{overlapping} experiment.\\

\textbf{Currency.}~ For this experiment, we use an initial number of $M=20$ inducing points. We choose a M\'atern kernel function with initial length-scale and noise amplitude values equal to $\ell = 10^{-3}$ and $\sigma_a = 0.1$, respectively. The incremental rule for the inducing-points is linear within time-steps. The VEM algorithm makes a maximum of 4 iterations per time-step.\\

\textbf{Banana.}~ In the two-dimensional input experiment for GP classification, we setup an initial grid of inducing-points with $M=3$ per side. The size of the grid increases within time as $M_t = M_{t-1} + 1$. In this case, we use an RBF kernel whose hyperparameters are initialized to $\ell=0.05$ and $\sigma_a = 0.1$. The maximum number of VEM iterations is fixed to 4 as well. For the binary prediction plots in Figure \ref{fig:banana}, we threshold the predictive probability as $p<0.5$ or $p\geq 0.5$ for $\yc_n = 1$, otherwise. The test-training data splitting is based on a $30\%$ proportion.\\

\textbf{Synchronous.}~ We generate $N=2000$ input-output samples where the output observation is multiple with $D=2$ real-valued dimension. As we consider a toy multi-task regression problem, we set a likelihood model that is defined using the syntax: \texttt{likelihoods\_list = [Gaussian(sigma=1.), Gaussian(sigma=2.0)]}, where we assume the Gaussian noise parameters $\sigma_n$ always fixed. We use $Q=2$ true latent functions $\Ucal$ that are defined by the expressions\\

\begin{equation*}
	u_1(x) = \frac{9}{2}\cos(2\pi x + \frac{3\pi}{2}) - 3\sin(4.3\pi x + \frac{3\pi}{10}) + 5\cos(7\pi x + 2.4\pi),
\end{equation*}
\begin{equation*}
u_2(x) = \frac{9}{2}\cos(\frac{3\pi}{2} x + \frac{\pi}{2}) + 5\sin(3\pi x + \frac{3\pi}{2}) -  \frac{11}{2}\cos(8\pi x +\frac{\pi}{4}), 
\end{equation*}

where the vectors $\bm{w}_q$ of the linear mixing are $\bm{w}_1 = [-0.5, 0.1]^{\top}$ and $\bm{w}_2 = [-0.1, 0.6]^{\top}$. Moreover, we choose an RBF kernel for the GP prior of both latent functions and their hyperparameters are initialized to $\ell=0.05$ and $\sigma_a = 0.5$. The number of inducing-points is $M_q = 5$ for both latent functions and increases linearly within time.

\textbf{Asynchronous.}~ The setup of this experiment is analogous to the synchronous case, where the slight difference is that the initial number of inducing-points per latent function $u_q$ is $M_q = 4$ instead. \\

\textbf{MOCAP.}~ For this experiment, we use a MOGP prior with $Q=3$ latent functions and an initial number $M_q = 10$ in all cases. The maximum number of VEM iterations is 5 in order to guarantee a good fitting. The multi-task likelihood model is defined by the syntax: \texttt{likelihoods\_list = [Gaussian(sigma=0.3), Gaussian(sigma=0.3), Gaussian(sigma=0.3)]}.\\

\textbf{Solar.}~ The solar dataset consists of a sequence of $t=1000$ real-valued observations. We use an extra batch with t=100 samples for a \textit{warm up} period. The initial number of inducing-points is $M=15$. We allow the VEM algorithm to make one iteration per continual update. The likelihood noise parameter is set to $\sigma_n = 1.0$. At each time-step, we initialize the RBF kernel of the GP prior to have a lengthscale $\ell=0.5$ and amplitude $\sigma_a = 2.0$. We only increase the number $M$ of inducing-points every 25 time-steps.\\

\vskip 0.2in
\bibliography{jmlr2019}

\end{document}